\newif\ifanonymous \anonymousfalse

\ifanonymous
    \documentclass[acmtog,anonymous,authorversion,nonacm]{acmart_arxiv/acmart}
    \else
    \documentclass[acmtog,authorversion,nonacm]{acmart_arxiv/acmart}
    \fi
\def\ShowNotes{}
\usepackage{cleveref}
\crefname{section}{Sec.}{Secs.}
\Crefname{section}{Section}{Sections}
\Crefname{table}{Table}{Tables}
\crefname{table}{Tab.}{Tabs.}
\Crefname{figure}{Figure}{Figures}
\crefname{figure}{Fig.}{Figs.}

\newcommand{\ignorethis}[1]{}

\newcommand{\etal       }     {{et~al.}}

\newcommand{\Reals      }     {{\textrm{I\kern-0.18em R}}}

\newcommand{\change     } [1] {\mbox{{\footnotesize $\Delta$} \kern-3pt}#1}

\definecolor{darkred}{rgb}{0.7,0.1,0.1}
\definecolor{darkgreen}{rgb}{0.1,0.6,0.1}
\definecolor{cyan}{rgb}{0.7,0.0,0.7}
\definecolor{otherblue}{rgb}{0.1,0.4,0.8}
\definecolor{maroon}{rgb}{0.76,.13,.28}
\definecolor{burntorange}{rgb}{0.81,.33,0}
\definecolor{olive}{RGB}{186, 184, 108}

\ifdefined\ShowNotes
  \newcommand{\colornote}[3]{{\color{#1}\textbf{#2} #3\normalfont}}
\else
  \newcommand{\colornote}[3]{}
\fi

\newcommand\todosilent[1]{}

\usepackage{booktabs} % For formal tables

\usepackage{graphicx}
\usepackage{adjustbox}
\usepackage{siunitx}
\usepackage[ruled]{algorithm2e} % For algorithms

\SetAlFnt{\small}
\SetAlCapFnt{\small}
\SetAlCapNameFnt{\small}
\SetAlCapHSkip{0pt}

\begin{document}
% Title portion
\title[V-LASIK]{V-LASIK: Consistent Glasses-Removal from Videos Using Synthetic Data}

\author{Rotem Shalev-Arkushin$^{1,2}$\hspace{1em} Aharon Azulay$^1$\hspace{1em} Tavi Halperin$^1$ \hspace{1em} Eitan Richardson$^1$ \\ 
\hspace{1em} \hspace{1em} Amit H. Bermano$^2$\hspace{1em}  Ohad Fried$^3$}

\affiliation{
 \institution{Lightricks$^1$ \hspace{0.3em} Tel Aviv University$^2$ \hspace{0.3em} Reichman University$^3$}
}

\affiliation{\url{https://v-lasik.github.io}}

\renewcommand\shortauthors{Shalev-Arkushin, R. et al.}
\begin{abstract}
Recently, diffusion-based generative models showed remarkable image and video editing capabilities. However, local video editing, particularly removal of small attributes like glasses, remains a challenge. Existing methods either alter the videos excessively, generate unrealistic artifacts, or fail to perform the requested edit consistently.
In this work, we focus on consistent and identity-preserving glasses-removal in videos, as a case study for video local attribute removal. We demonstrate the generalizability of our method by applying it to facial sticker removal from videos.
Due to the lack of paired data, we adopt a weakly supervised approach and generate synthetic imperfect data, using an adjusted pretrained diffusion model. We show that despite data imperfection, by learning from our generated data and leveraging the prior of pretrained models,
our model is able to perform the desired edit consistently while preserving the original video content. 
Furthermore, we suggest a new normalization method, Inside-Out Normalization, which aligns colors in a filled region with colors outside that region.
Our approach offers significant improvement over existing methods, showcasing the potential of leveraging synthetic data and strong priors for local video editing tasks.
\end{abstract}

\begin{teaserfigure}
  \centering
   \includegraphics[width=\textwidth]{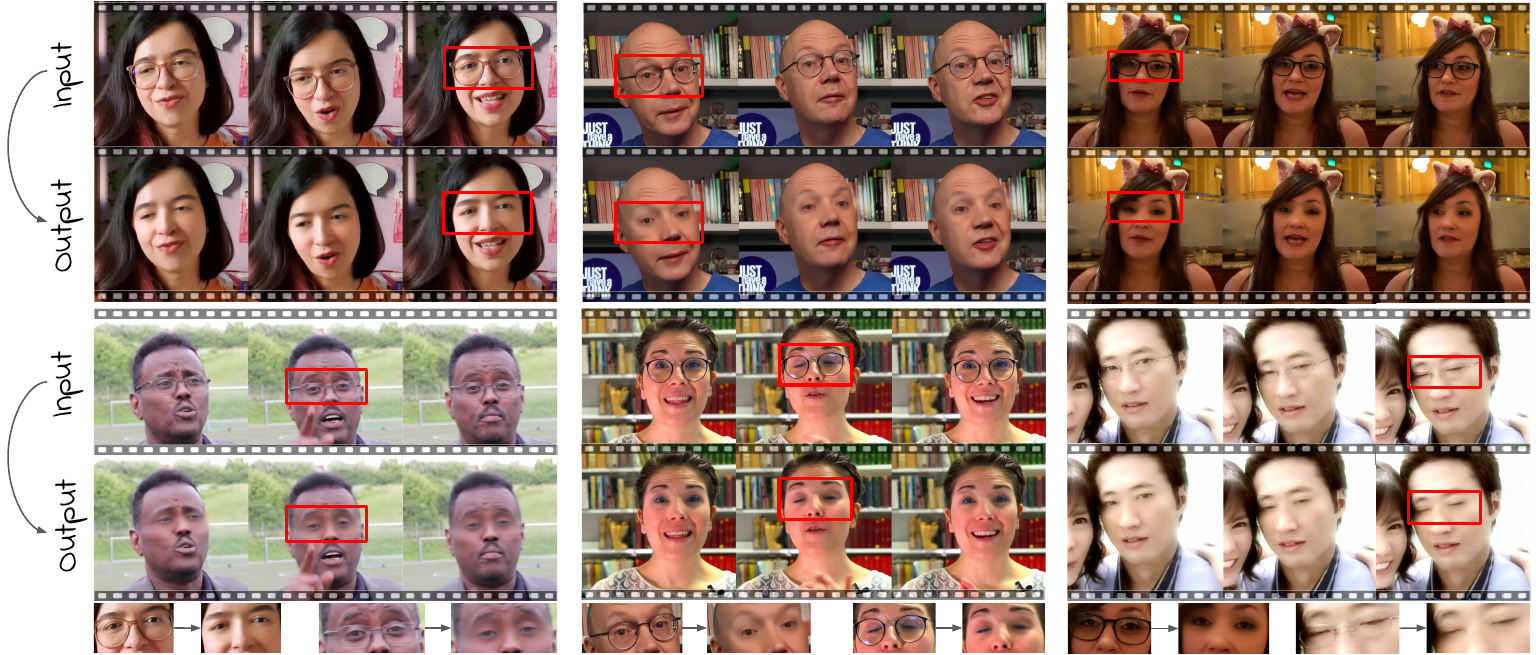}
   \caption{Our method receives an input video of a person wearing glasses, and consistently removes the glasses, while preserving the original content and identity of that person. Our method successfully removes the glasses even when there are reflections (bottom-middle example), heavy makeup (top-right), and eye blinks (bottom-right). Red rectangles are zoomed-in at the bottom row.
   %The last row presents a zoom-in to the eyes marked with red squares at each example.
   }
   \label{fig:teaser}
   \Description[]{}  % use this line to please the compiler
\end{teaserfigure}

\maketitle

% Manuscript Body
\section{Introduction}
\label{sec:intro}

\begin{figure}[t]
    \centering
    \setlength{\tabcolsep}{1pt}
    \renewcommand{\arraystretch}{0.5}

\begin{adjustbox}{max width=\linewidth}
    \begin{tabular}{c@{\hspace{0.5em}}ccccccc}

        Input &
        LEDITS &
        IP2P &
        Lyu \etal &
        SD inpaint &
        CN inpaint &
        Synth data &
        \textbf{Ours} \\

        \includegraphics[clip,width=18mm]{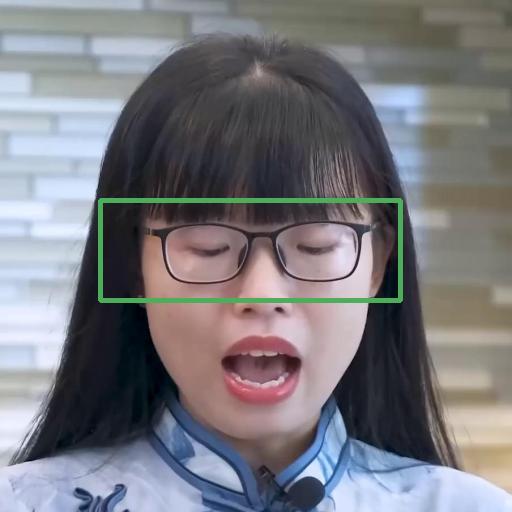} &
        \includegraphics[clip,width=18mm]{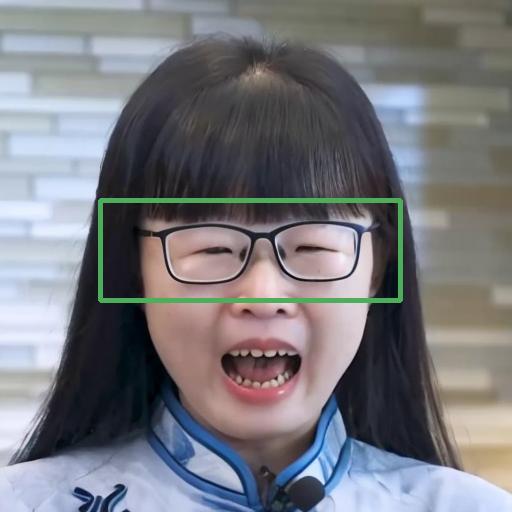} &
        \includegraphics[clip,width=18mm]{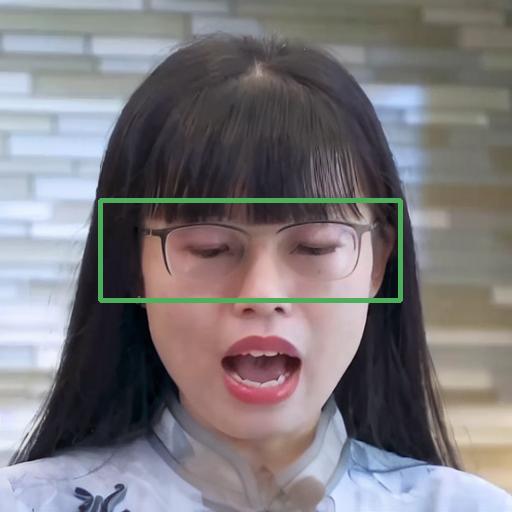} &
        \includegraphics[clip,width=18mm]{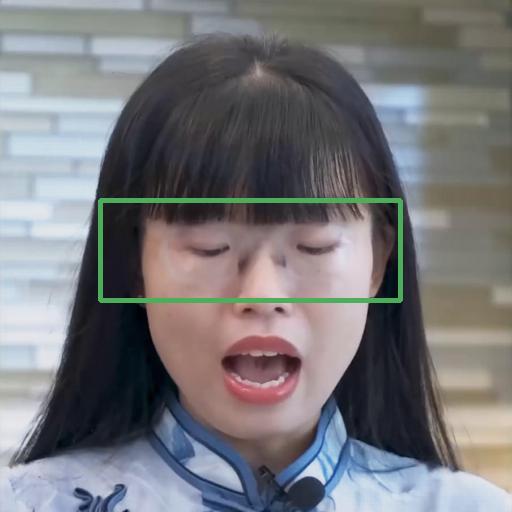} &
        \includegraphics[clip,width=18mm]{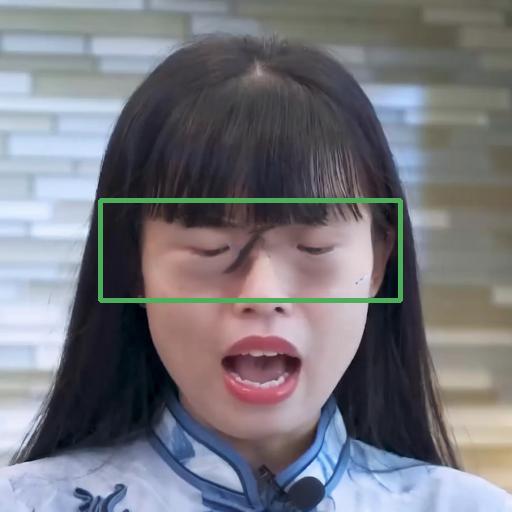} &
        \includegraphics[clip,width=18mm]{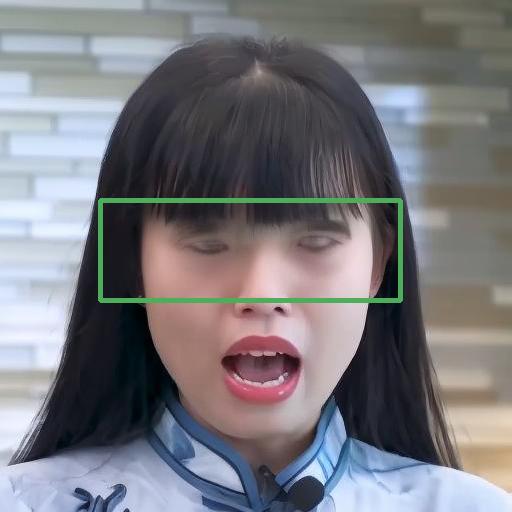} &
        \includegraphics[clip,width=18mm]{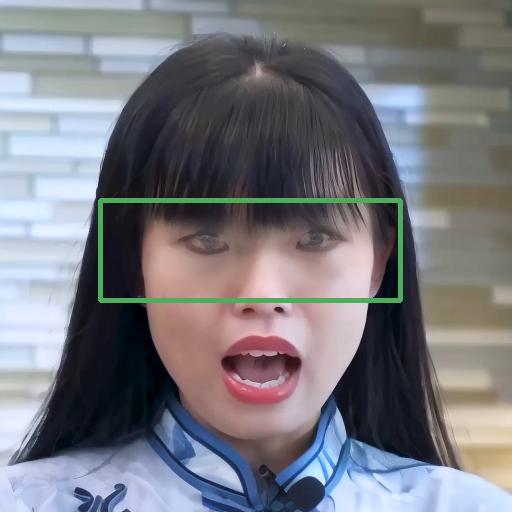} &
        \includegraphics[clip,width=18mm]{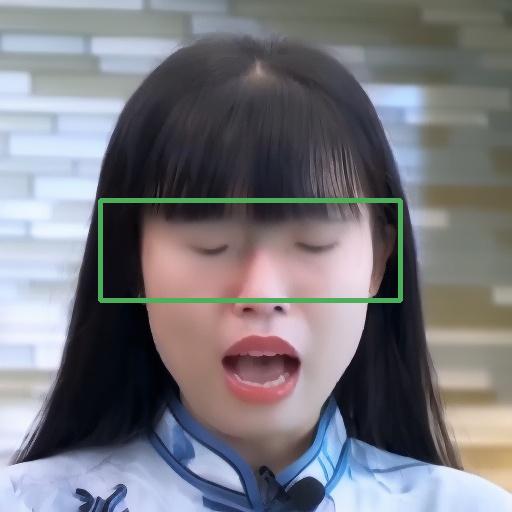} 
        \\

        \includegraphics[clip,width=18mm]{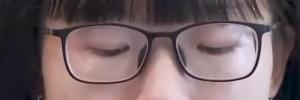} &
        \includegraphics[clip,width=18mm]{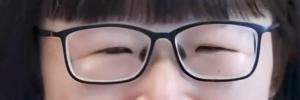} &
        \includegraphics[clip,width=18mm]{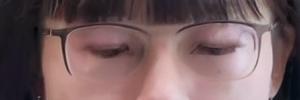} &
        \includegraphics[clip,width=18mm]{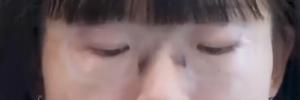} &
        \includegraphics[clip,width=18mm]{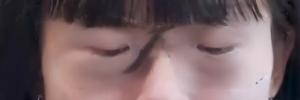} &
        \includegraphics[clip,width=18mm]{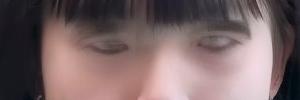} &
        \includegraphics[clip,width=18mm]{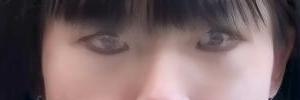} &
        \includegraphics[clip,width=18mm]{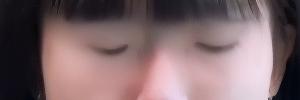} 

    \end{tabular}
\end{adjustbox}
    \caption{
    \textbf{Glasses-removal from a blinking eye by image editing methods}
    Left to right: LEDITS~\shortcite{tsaban2023ledits}, Instruct pix2pix~\shortcite{brooks2023instructpix2pix}, Lyu \etal~\shortcite{lyu2022portrait}, Stable Diffusion inpaint~\shortcite{rombach2022high}, ControlNet inpaint~\shortcite{zhang2023adding}, our synthetic dataset generation result, and our final result. As image editing methods expect high quality images with people looking straight to the camera, they struggle when these constraints are not met. In our dataset result `Synth data', as a result of the cross-frame attention, eye artifacts appear. However, our model is still able to learn from the imperfect data and remove the glasses better than any out-of-the-box method, and better than the data it was trained on.}
    \vspace{-5pt}
    \label{fig:im_edit}
       \Description[]{}  % use this line to please the compiler
       
\end{figure}

Recent advances in diffusion-based generative models \citet{ho2020denoising, sohl2015deep} have demonstrated impressive capabilities in image and video editing \citep{rombach2022high, zhang2023adding, geyer2023tokenflow, yang2023rerender, zhang2023controlvideo, kara2023rave, bar2024lumiere}. 
While local editing, such as removing attributes without changing the rest of the content, mainly works for images, for videos it remains a challenge. Video frames often contain motion blur and challenging poses that are less common in images. Moreover, videos of people pose additional challenges, as humans are highly sensitive to subtle unrealistic ``uncanny'' artifacts, and our visual system is more sensitive to motion inconsistencies, which leads to a higher quality bar.
Existing image editing and inpainting methods \citet{meng2021sdedit, zhang2023adding, tsaban2023ledits, brooks2023instructpix2pix, couairon2023diffedit, nitzan2024lazy} are trained on images that do not contain the aforementioned problems, and therefore do not perform well over video frames; They either change the frames too much, or do not perform the requested edit completely, as illustrated in \cref{fig:im_edit}.
Video inpainting methods \citet{zhou2023propainter, yu2023inpaint, li2022towards, zhang2022flow} typically remove objects by deleting and filling regions of the video, often disregarding the information originally present in those regions. When trying to generate new video content behind an object that never moves, these methods tend to generate a low detailed background, or one with clear unrealistic artifacts. 
As a representing use-case for local attribute removal, we choose to tackle glasses removal, as it is a particularly challenging example; where some content is never or only partially revealed (e.g. eyes, eyebrows), and is changed across the video (e.g. eyelids change position). 
Since eyes contribute greatly to identity perception, a consistent and highly realistic generation is required for an identity to be preserved. However, current video inpainting works typically generate smooth results when using small masks, that do not fit the facial case well (see ProPainter \citet{zhou2023propainter}, FGT \citet{zhang2022flow} results in \cref{fig:results_comp}).
Mask-free video editing methods \citet{geyer2023tokenflow, zhang2023controlvideo, khachatryan2023text2video, kara2023rave, cong2024flatten} mainly focus on changing the style of the whole video or large parts of it, and not on local changes that should adhere to the original context.
Other works \citet{kasten2021layered, bar2022text2live} learn an atlas for each new video, leading to long inference times. Furthermore, the goal of these works is to manipulate the appearance of existing objects, not to change the composition of the elements in the scene \citep{bar2022text2live}. Hence, their method is not capable of successfully removing attributes such as glasses from faces.

In this work, we introduce a new method that consistently removes glasses from videos while preserving the content of the original video. To the best of our knowledge, no paired data exists for this task; Therefore, we cannot finetune a model directly to solve it. To tackle this problem, we generate imperfect synthetic data, and use it to train a model in a weakly supervised manner. We show that by using an adjusted pretrained image inpainting model, we can generate imperfect data pairs, and learn from them. Our method is closest to video inpainting in the sense that it removes something from the video and replaces it with different content. However, similarly to recent video editing works, the input to our model is a mask-free image, exposing it to facial details that are occluded for inpainting models.
As glasses often contain colored lens and reflections that change across frames, which affect the way the eye behind them looks, they are a challenging attribute to remove from faces. Hence, we incorporate cross-frame attention layers in our data generation model, which allows aggregating information from different frames.
Still, as presented in \cref{fig:im_edit}, the generated data (`Synth data') is imperfect and contains many artifacts, such as not always preserving the eyelids positions from the original frame. To overcome these challenges, we finetune a pretrained diffusion based image-to-image model over this data. This way, we can leverage its prior, while enhancing its capabilities with our new data. 
Despite imperfections in our synthetic data, the model can leverage its prior knowledge to outperform the training data and produce high-quality results.
For temporally consistent results, we combine our trained model with a motion prior. Finally, for smooth and content-faithful edits, we suggest a novel normalization method, Inside-Out Normalization (ION) that allows matching the colors of a masked region with the area outside the mask. We achieve state-of-the-art results on video glasses-removal, surpassing all current video editing and inpainting methods.
To demonstrate that our method can be generalized to other local video editing tasks, we successfully apply it to the task of removing stickers from faces.
\section{Related Work}
\label{sec:prior_work}

\subsection{Image editing}
Since the advent of diffusion models for image generation and editing \citet{ho2020denoising}, real image editing has rapidly advanced and showed remarkable results \citet{brooks2023instructpix2pix, couairon2023diffedit, tsaban2023ledits, hertz2022prompt, tumanyan2023plug, zhang2023adding, meng2021sdedit}, including local editing such as glasses-removal. 
Image inpainting methods \citet{nitzan2024lazy, rombach2022high, zhang2023adding, yildirim2023inst}, also showed impressive results, realistically replacing content behind a given mask.
Moreover, prior work \citet{lyu2022portrait, lee2020byeglassesgan} tackled the task of image glasses-removal directly. These works perform well over images. However, as shown in \cref{fig:im_edit} they fail when applied to video frames of people, that do not always look directly at the camera, and constantly move, causing motion blur and other artifacts.
Moreover, when applied frame-by-frame, the generated content differs between frames, resulting in temporal inconsistencies.
Prior to diffusion models, Generative Adversarial Networks (GANs) have been widely used for facial image editing \citep{karras2019style, karras2020analyzing, yao2021latent, patashnik2021styleclip, richardson2021encoding}. However, In addition to temporal consistency issues, they operate on aligned crops of faces, thus when applied to videos, faces must be stitched back to the original frames which is challenging and often yields unsatisfactory results.

\subsection{Video editing}
Recently, video editing has developed greatly with video-to-video translation methods \citet{geyer2023tokenflow, khachatryan2023text2video, cong2024flatten, kara2023rave, qi2023fatezero, wu2023tune, zhang2023controlvideo} that focus on transforming the entire frame into a different style, while trying to preserve temporal consistency in the generated videos. Some of them, such as RAVE~\citet{kara2023rave}, or TokenFlow~\citet{geyer2023tokenflow} with SDEdit~\citet{meng2021sdedit} also perform local editing. However, in the case of glasses-removal, where a person moves across the video and the requested edit is small and delicate, many artifacts such as face deformations and inconsistencies are generated. Examples are presented in \cref{fig:results_comp} and in video results in the supplementary material. 
Earlier work used GANs to perform local editing of faces \citep{tzaban2022stitch, xu2022temporally, xu2023rigid}.
Another line of works is atlas-based video editing \citep{kasten2021layered, lu2021omnimatte, bar2022text2live, suhail2023omnimatte3d}. These methods learn an atlas for each video, to apply changes to either the background or foreground of the video. The initial atlas reconstruction requires excessive computational resources and long running times. Moreover, as these methods were designed to change the appearance of existing objects, they are not meant for adding or removing attributes that were not part of the original video, such as glasses, and do not perform these changes well. 

\subsection{Video inpainting}
Video inpainting is the task of consistently filling-in new content behind a given mask throughout a video \citep{yu2023inpaint, zhang2022flow, zhou2023propainter, gu2023flow, li2022towards}. Such methods work well when the object moves throughout the video, and the model is able to fill-in the background from other video frames, where the background is visible. However, they struggle with filling-in completely occluded areas, especially when they are part of the foreground, such as faces. Particularly, areas behind glasses are only partially revealed by head motion, and sometimes glasses cause optical deformations that hide the background. Slight changes to face features in the ``background'' of glasses may yield unrealistic results, or alter face identity, as shown in the ProPainter~\citet{zhou2023propainter} and FGT~\citet{zhang2022flow} results in \cref{fig:results_comp}. While such changes to smooth backgrounds may not affect the quality of the results significantly, in foreground areas, particularly in faces, they are detrimental.

\subsection{Learning from synthesized data}
Deep learning models require large amount of data to learn from, however paired data is not always available for the task at hand. Hence, several works \citet{li2022bigdatasetgan, peebles2022gan, ravuri2019classification, brooks2023instructpix2pix, lyu2022portrait} used generative models for data generation to train models. These works usually rely on the high quality of the generated data, and learn solely from it. In this work, we accept that generated data is usually imperfect, and is insufficient to achieve the required result on its own, thus we take advantage of strong priors of pretrained models and use our generated data for fine-tuning. This way, the model learns to generate results that are superior to the data it was trained on, while performing the relevant task. 
\citet{lyu2022portrait} specifically explores synthetic data for the task of glasses-removal from images. However, their data acquisition process requires face scanning and 3D data, which is hard to acquire, unlike our method which does not require any special equipment or effort.
\section{Method}
\label{sec:method}
\begin{figure}[t]
    \centering
    \setlength{\tabcolsep}{2pt}
    \renewcommand{\arraystretch}{0.5}

\begin{adjustbox}{max width=\linewidth}
    \begin{tabular}{cccccccc}

        Input &
        Masked input &
        w/o CF attn &
        w/ CF attn &
        Input &
        Masked input &
        w/o CF attn &
        w/ CF attn
        \\

        \includegraphics[clip,trim=0 9.5 0 10,width=23mm]{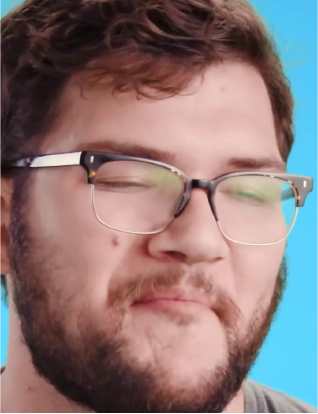} &
        \includegraphics[clip,trim=0 11 0 10,width=23mm]{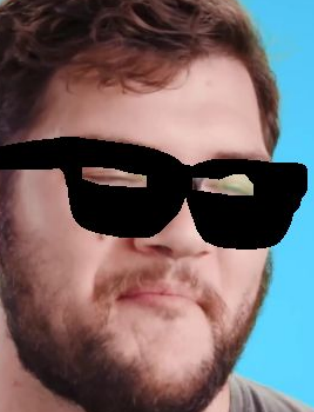} &
        \includegraphics[clip,trim=0 12 0 10,width=23mm]{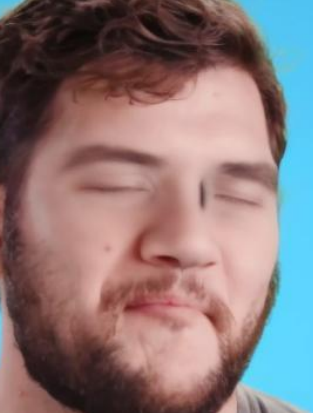} &
        \includegraphics[clip,trim=0 8.5 0 10,width=23mm]{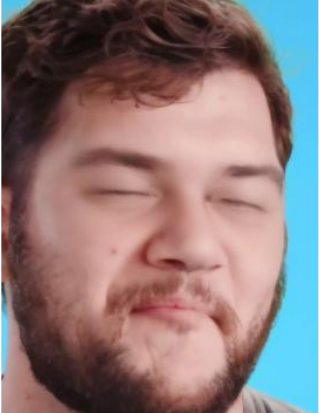} &
        \includegraphics[clip,trim=0 0 0 15,width=23mm]{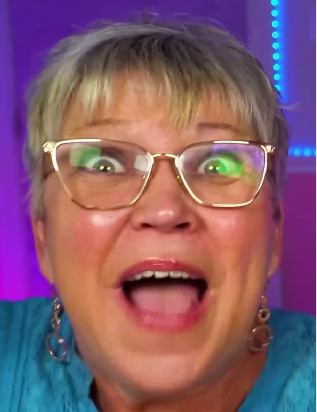} &
        \includegraphics[clip,trim=0 2.6 0 15,width=23mm]{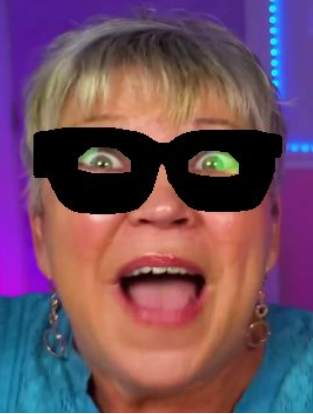} &
        \includegraphics[clip,trim=0 0.4 0 15,width=23mm]{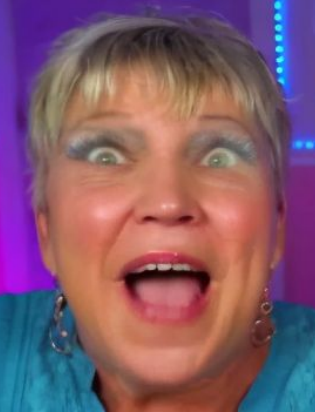} &
        \includegraphics[clip,trim=0 0.9 0 15,width=23mm]{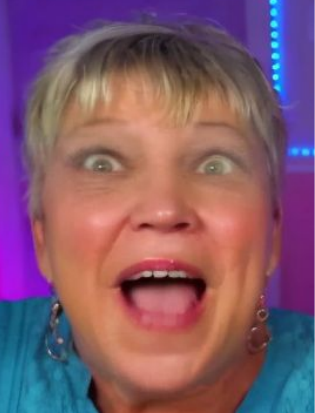}
        \\
        \includegraphics[clip,trim=25 0 15 0,width=23mm]{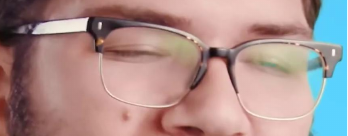} &
        \includegraphics[clip,trim=25 0 15 0,width=23mm]{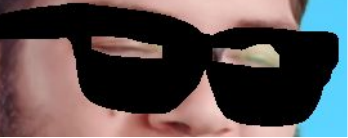} &
        \includegraphics[clip,trim=25 0 15 0,width=23mm]{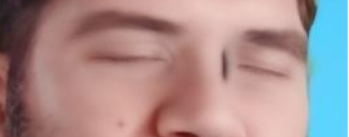} &
        \includegraphics[clip,trim=25 0 15 0,width=23mm]{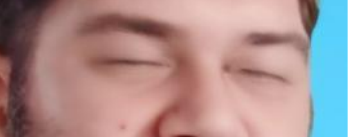} &
        \includegraphics[clip,trim=23 0 23 0,width=23mm]{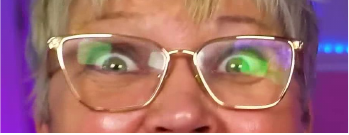} &
        \includegraphics[clip,trim=23 0 23 0,width=23mm]{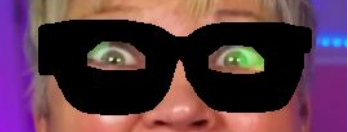} &
        \includegraphics[clip,trim=23 0 23 0,width=23mm]{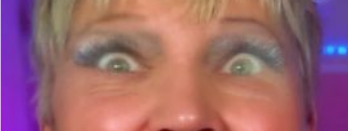} &
        \includegraphics[clip,trim=23 0 23 0,width=23mm]{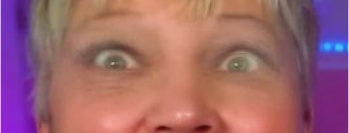}
        \\
        input & inaccurate mask & glasses remnants & no remnants & input & masked input & reflections & no reflections

    \end{tabular}
\end{adjustbox}
    \caption{
    \textbf{Cross-frame attention importance in data generation.} Cross-frame attention helps removing glasses remnants, even when the mask is not perfect (left example) and reducing glasses reflections (right example).
    }
    \vspace{-5pt}
    \label{fig:ablation}
       \Description[]{}  % use this line to please the compiler
\end{figure}
\begin{figure*}[t]
  \centering
  % \adjustbox{trim=0cm 0cm 0cm 2.2cm}{
   % \includesvg[width=\textwidth]{Assets/overview.svg}
   % }
   \includegraphics[width=\linewidth]{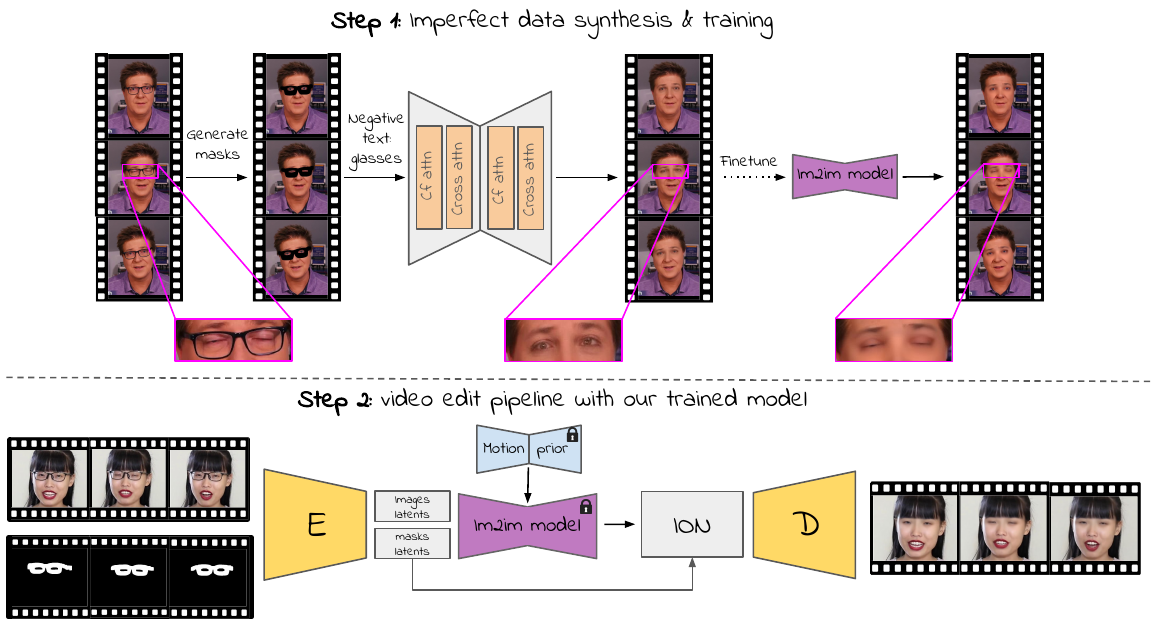}
   \caption{\textbf{Method overview:} \textbf{Step 1:} we create an imperfect synthetic paired dataset by generating glasses masks for each video frame and inpainting it. We inpaint each frame using an adjusted version ControlNet inpaint \cite{zhang2023adding}. We replace the self-attention layers with cross-frame attention (cf attn) and use blending between the generated latent images and the noised masked original latent images at each diffusion step. The generated data in the first step is imperfect; e.g. in the middle frame, the person blinks, however its generated pair has open eyes. Nevertheless, the data is good enough for finetuning an image-to-image diffusion model and achieving satisfactory results, due to the strong prior of the model. 
   \textbf{Step 2:} Given our trained model for the task of removing glasses from images, we incorporate it with a motion prior module to generate temporally consistent videos without glasses from previously unseen videos. To obtain the original frame colors, at each diffusion step we blend the generated frames with the noised original masked latent images, and before decoding, we apply an Inside-Out Normalization (ION), to better align the statistics within the masked area and the area outside of the mask.
}
   \label{fig:overview}
      \Description[]{}  % use this line to please the compiler
\end{figure*}

Given an input video of a person wearing glasses, our goal is to remove the glasses while preserving all other information.
As illustrated in \cref{fig:overview}, our approach to removing glasses from videos consists of three stages: data generation, training, and editing.
First, as no paired data is available for this task, we generate a synthetic paired dataset using videos of people wearing glasses. 
Next, we finetune an image-to-image diffusion model over our synthetic dataset, to get realistic video frames without glasses, where all other parts of the frame remain similar to the original frame, and the identity of the edited person is preserved.
Finally, we incorporate our trained model with a pretrained motion module, into a video editing pipeline, to obtain temporally consistent videos without glasses.

\subsection{Paired Data Generation}
\label{subsec:data}

As illustrated in \cref{fig:im_edit}, current image editing and inpainting methods do not perform well on the task of removing glasses from video frames. Therefore, we wish to train a model for this task. However, to the best of our knowledge, no relevant paired dataset currently exists. Hence, we create a synthetic dataset of paired video frames, with and without glasses, using videos from the CelebV-Text dataset~\cite{yu2023celebv} where people wear glasses. 
For each video frame, we generate a glasses mask using a face parser~\cite{zheng2022farl} that identifies the glasses in it. The eyes are a key component in the identity of a person, and the current position of the eyelids is crucial to generating a result that is consistent with the original video, that might contain eye-closure and blinking.
Therefore, we want to give the model information about the eyes and eyelids, to keep their original appearance. To do that, we make eye ``holes'' in the glasses masks, using the identified eye landmarks. Then, we inpaint the glasses area by applying an adjusted inpainting diffusion model over each video frame and its generated mask. 

We make two adjustments to the inpainting model: First, to achieve smoother and more realistic transitions between the frame and the inpainted part, we blend the latent feature vectors with a noised encoded version of the masked original frame at each diffusion step, as suggested in Blended~Latent~Diffusion~\cite{avrahami2023blended}.
Second, to increase the edit consistency across frames from the same video, we replace the self-attention layers of the model with a cross-frame attention \cite{khachatryan2023text2video}.
For each video frame, we perform a cross-frame attention with $k$ reference frames from the same video. We use multiple reference frames as sometimes information such as the eyebrows or the eye color is hidden behind the glasses or their reflections. Using cross-frame attention with multiple frames allows for generalization from all $k$ frames. 
For this purpose, we adopt the cross-frame attention mechanism suggested by TokenFlow~\cite{geyer2023tokenflow}: 
\begin{equation}
\label{eq_cf_attn}
\text{softmax}(\frac{Q\cdot [K_0,...,K_k]^T}{\sqrt{d}})\cdot [V_1,...,V_k]
\end{equation}
where each query frame ($Q$) attends to $k$ different values ($V_i$), corresponding to $k$ different reference frames in our case.
As a result, even when face parts are occluded, the information about the occluded parts can be retrieved from the reference frames if they are revealed there. 
An example of occluding reflections is presented on the right side of \cref{fig:ablation}. In this example, there are reflections over the glasses, hiding the true eye colors behind them. Without using cross-frame attention, reflections and artifacts are present on the output result. However, when using cross-frame attention, the true eye color is preserved and no reflections are shown in the result.
Moreover, using cross-frame attention also helps when the masks are not exact and do not cover all of the glasses in all video frames, as in the left example of \cref{fig:ablation}. In this case, glasses remnants appear when no cross-frame attention is used, due to an incomplete mask. However, they are removed when we use cross-frame attention, where the masks of the reference frames cover the uncovered part. 

\subsection{Model fine tuning}
\label{subsec:finetune}
After the data generation stage, we obtain pairs of frames with and without glasses. However, our generated frames are not temporally consistent, and also contain per-frame artifacts in many cases, as featured in \cref{fig:im_edit,fig:overview}. For example, the eyelid positions, i.e. closed or open, sometimes change after the inpainting process. 
Despite that, the generated data is good enough for finetuning a pretrained image-to-image diffusion model for the task of removing glasses from faces. 
 
These models have strong prior for generating realistic looking images that are similar to the original ones and preserve their small and delicate details. Therefore, after finetuning, the model learns the task of removing glasses from our data, while preserving fine details such as eye color and eyelid positions. 

\subsection{Video editing pipeline}
\label{subsec:vid_edit}

To consistently remove glasses from unseen videos, we integrate our trained model with a pretrained motion module \cite{guo2023animatediff}. 
When applied directly to video frames, this model's results often exhibit slightly different colors from the original input. To overcome this issue, we can use the masks from the data generation stage to retrieve the original values on areas that are outside of the mask. For a smooth result, at each diffusion step we perform a gradual blending between the noised masked input and the generated result, such that the area within the mask changes completely, and the areas that surround the mask change less, gradually decreasing the amount of change as the pixels are further from the mask.
Additionally, inspired by AdaIN \cite{huang2017arbitrary}, we apply a new normalization function, we dub Inside-Out Normalization (ION), where we aim to align the statistics of the masked area with those of the non-masked area.
Formally, we calculate the mean and standard deviation of the masked area and the non-masked area, $\mu_{m}, \sigma_{m}$ and $\mu_{\bar{m}}, \sigma_{\bar{m}}$ respectively, and we normalize the values of the latent features in the area inside the mask by calculating:
\begin{equation}
    ION(x) = \sigma_{\bar{m}}\frac{x-\mu_{m}}{\sigma_{m}}+\mu_{\bar{m}}
\end{equation}

ION allows a smooth transition between the areas inside and outside of the mask by moving the statistics of the latent masked area toward those of the non-masked area. 

\section{Implementation Details}
\label{sec:imp}

\textbf{Data generation \& training specifics.}
For the glasses masks generation, we use Facer~\cite{zheng2022farl} 
with \texttt{retinaface/mobilenet} as a face detector, \texttt{farl/celebm/448} as a face parser, and 
\texttt{farl/\-ibug300w/448} as a face aligner. We first find the glasses mask using the parser, and then make eye holes in it based on the eye landmarks found by the face aligner. To generate the holes, for each eye, we connect the eye landmarks into one connected component, dilate it with a (10,10) kernel, so that we keep enough of the eyes information, and remove the component from the mask. Additionally, after resizing the mask to match the size of the latent vectors, we dilate the mask with a kernel of (3,3), and then blur it with a (3,3) kernel as well, to make sure we include all glasses pixels, and not too much from the rest of the image.
For the inpainting process, we use CN inpaint \cite{zhang2023adding} as our model, with latent blending with a blending ratio of 0.9.
We use 2 reference frames for the cross-frame attention: the first and middle frames in each video, since usually the person moves and changes position throughout the video so more information is gathered this way. If more memory is available, more reference frames will probably give better results. Moreover, for consistency we also use the same noise (encoding of the first reference frame) for all the video.
We use CN Tile \cite{zhang2023adding} as the model we finetune over our dataset, with batch size 8, learning rate \num{1e-5}. To avoid learning the artifacts of our imperfect data, and avoid forgetting the prior knowledge of CN Tile, we stop the training at an early stage, as suggested by DVP \cite{lei2020blind}.

\textbf{Editing pipeline specifics.}
We use the pretrained motion module of AnimateDiff \cite{guo2023animatediff} with context length 16, context overlap 4 as our motion prior module.
As shown in the right example of \cref{fig:lim}, the motion module tends to smooth the frames to get a more temporally consistent result, hence sometimes the results using this module get blurry. To avoid blurriness, we use only some of the motion layers and not all of them. Specifically, we remove the first 5 output motion layers.
This way, we get a more realistic and less blurry video.
For blending we use gradual values between 0 and 0.7 as mask values, so that pixels outside of the mask can also change a little bit for a smoother result. 

\textbf{Original colors vs. glasses removal trade-off.}
As an option in our video editing pipeline, we blend original pixels back into areas outside the masked region. If we do not blend them back, the colors in the edited video may not match those of the original video exactly. However, as the masks are not perfect, glasses remnants are sometimes left within the non-masked region, causing glasses remnants to appear in the result when using masks. To avoid it, we use dilation and different blending values outside of the mask. If the original colors are not as important and glasses-removal is of higher priority, one can use higher dilation for the mask, lower mask blending values, or even not use the masks at all.

\textbf{General.}
We use Stable Diffusion 1.5 \cite{rombach2022high} 
as our backbone in all the method steps, as at the time of development there was no compatible motion module for SDXL.
\section{Experiments}
\label{sec:eval}

\begin{figure*}[htpb]
    \centering
    \setlength{\tabcolsep}{0.1pt}

    \resizebox{.86\textwidth}{!}{
    \begin{tabular}{c@{\hskip 0.5em} c c c@{\hskip 0.5em} c c c}

        \raisebox{0.235in}{\rotatebox[origin=t]{90}{Input}} &
        \includegraphics[clip,width=16mm]{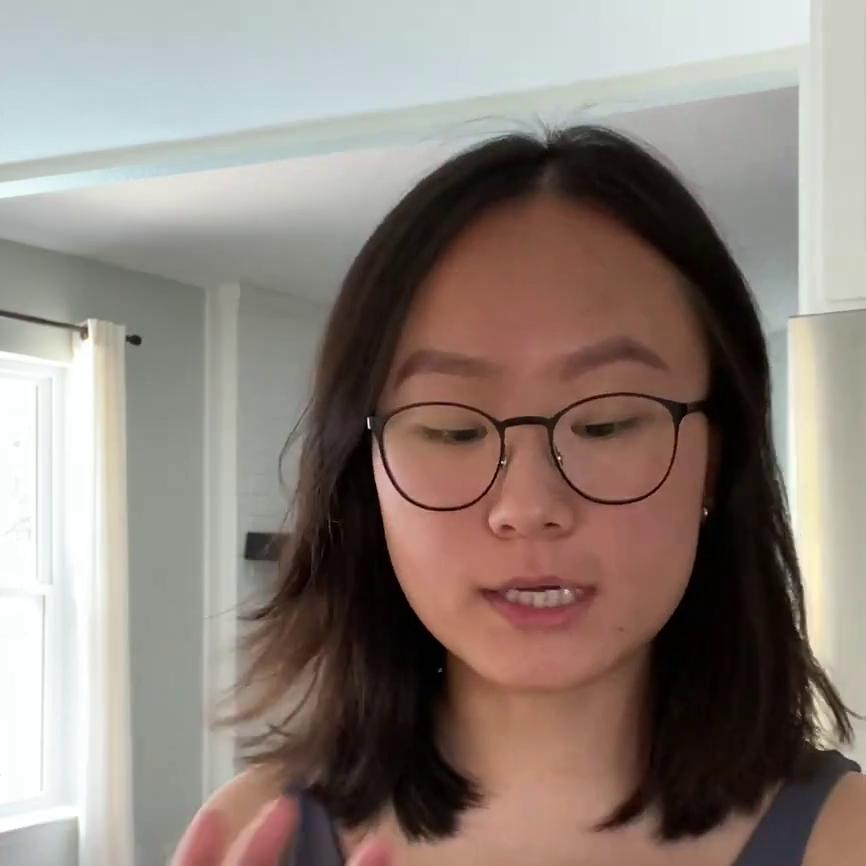} &
        \includegraphics[clip,width=16mm]{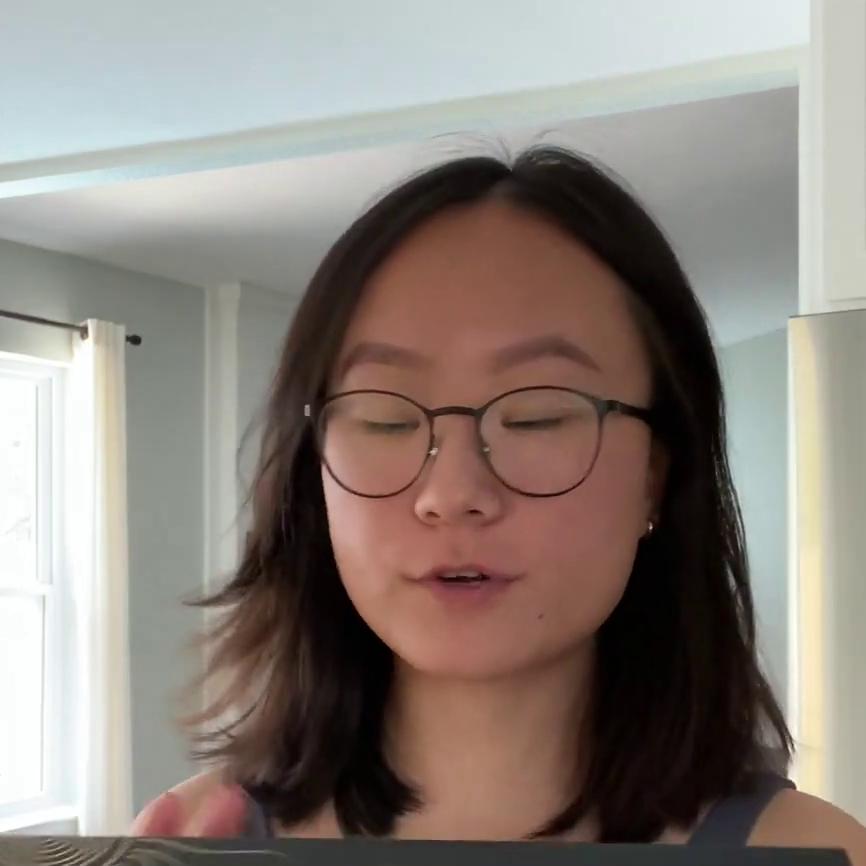} &
        \includegraphics[clip,width=16mm]{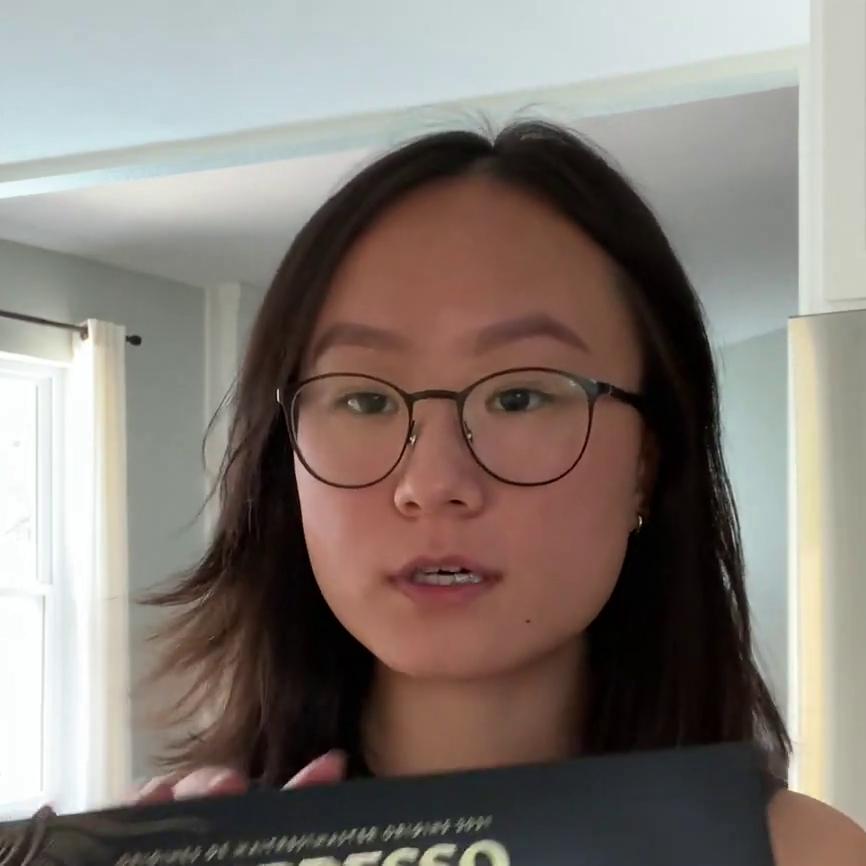} &
        \includegraphics[clip,width=16mm]{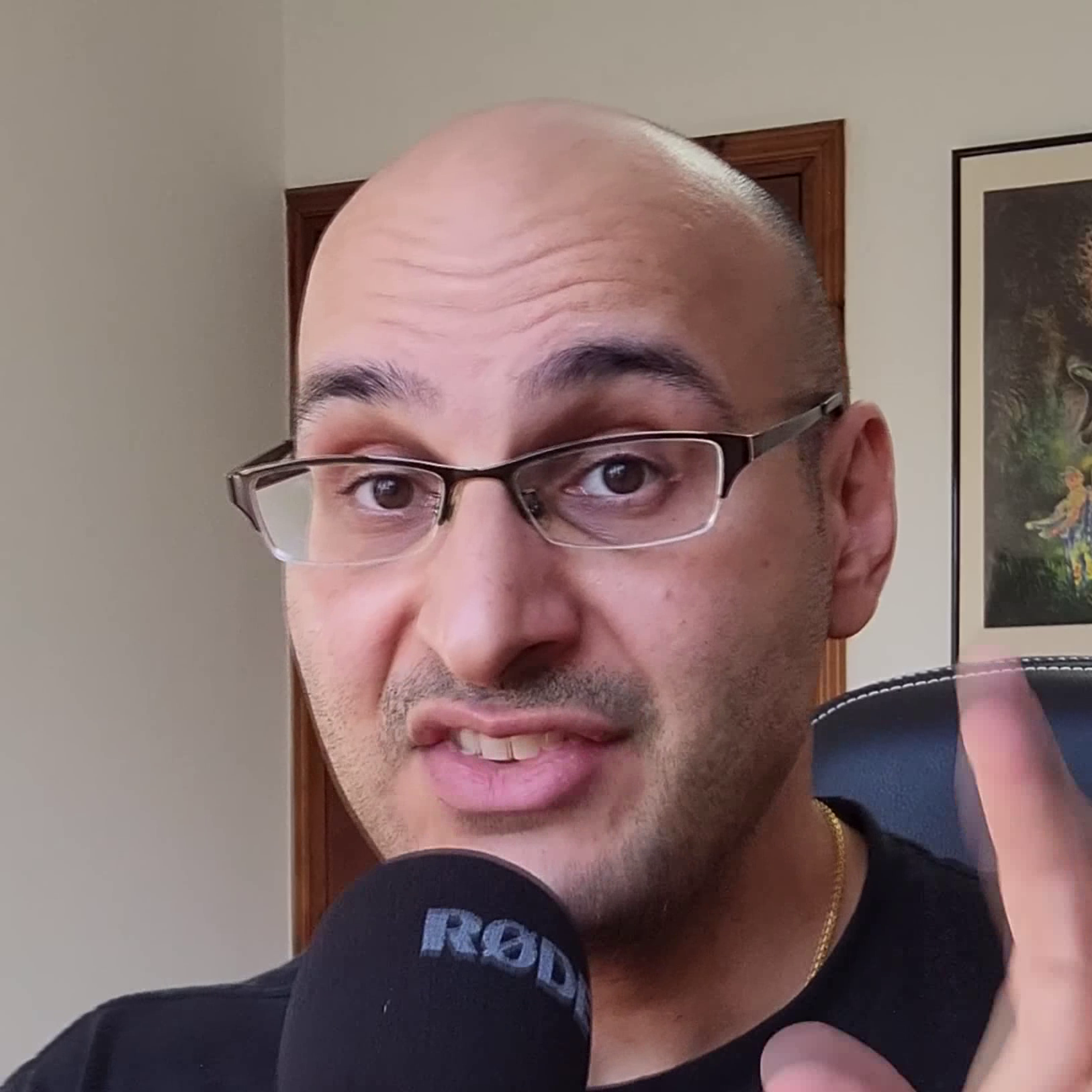} &
        \includegraphics[clip,width=16mm]{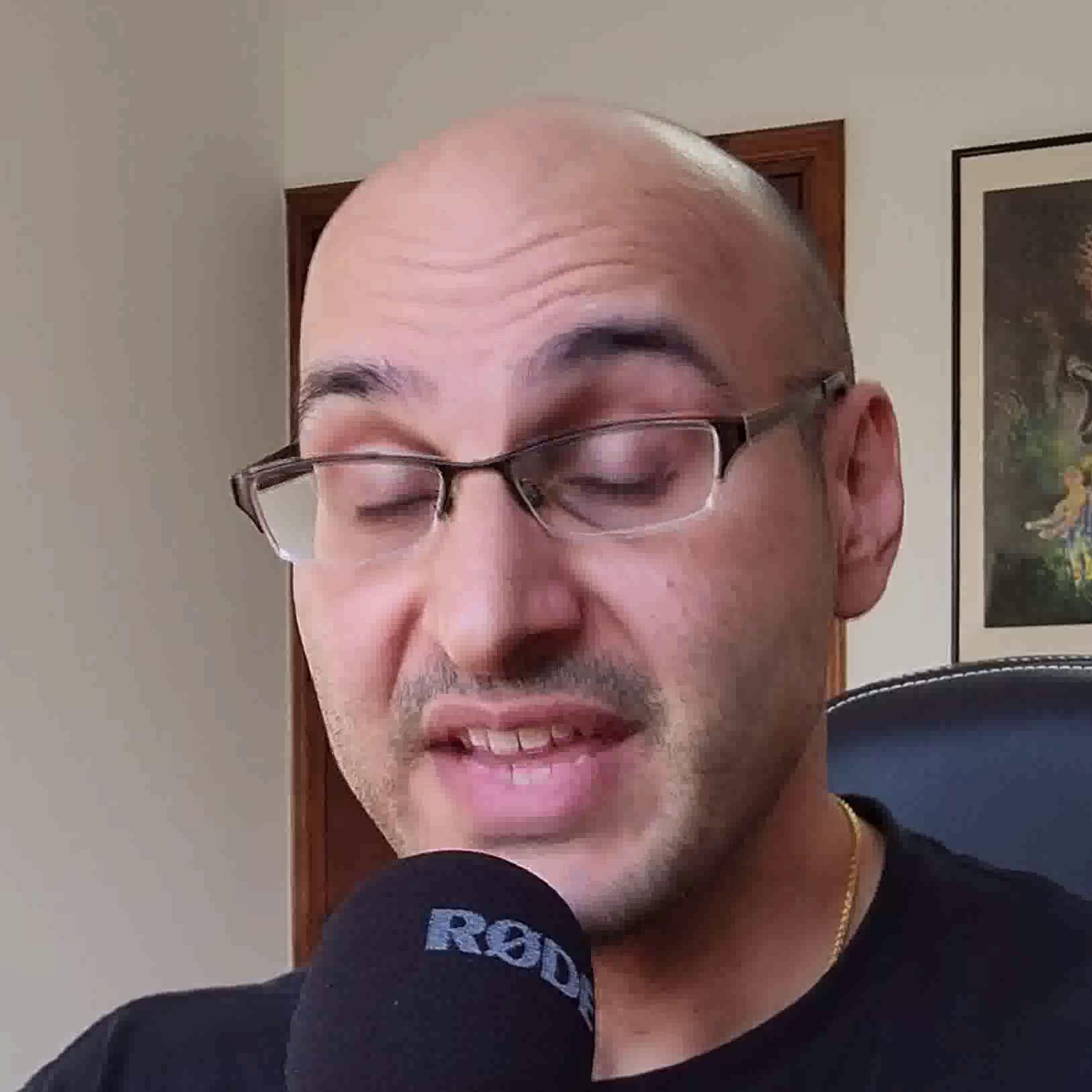} &
        \includegraphics[clip,width=16mm]{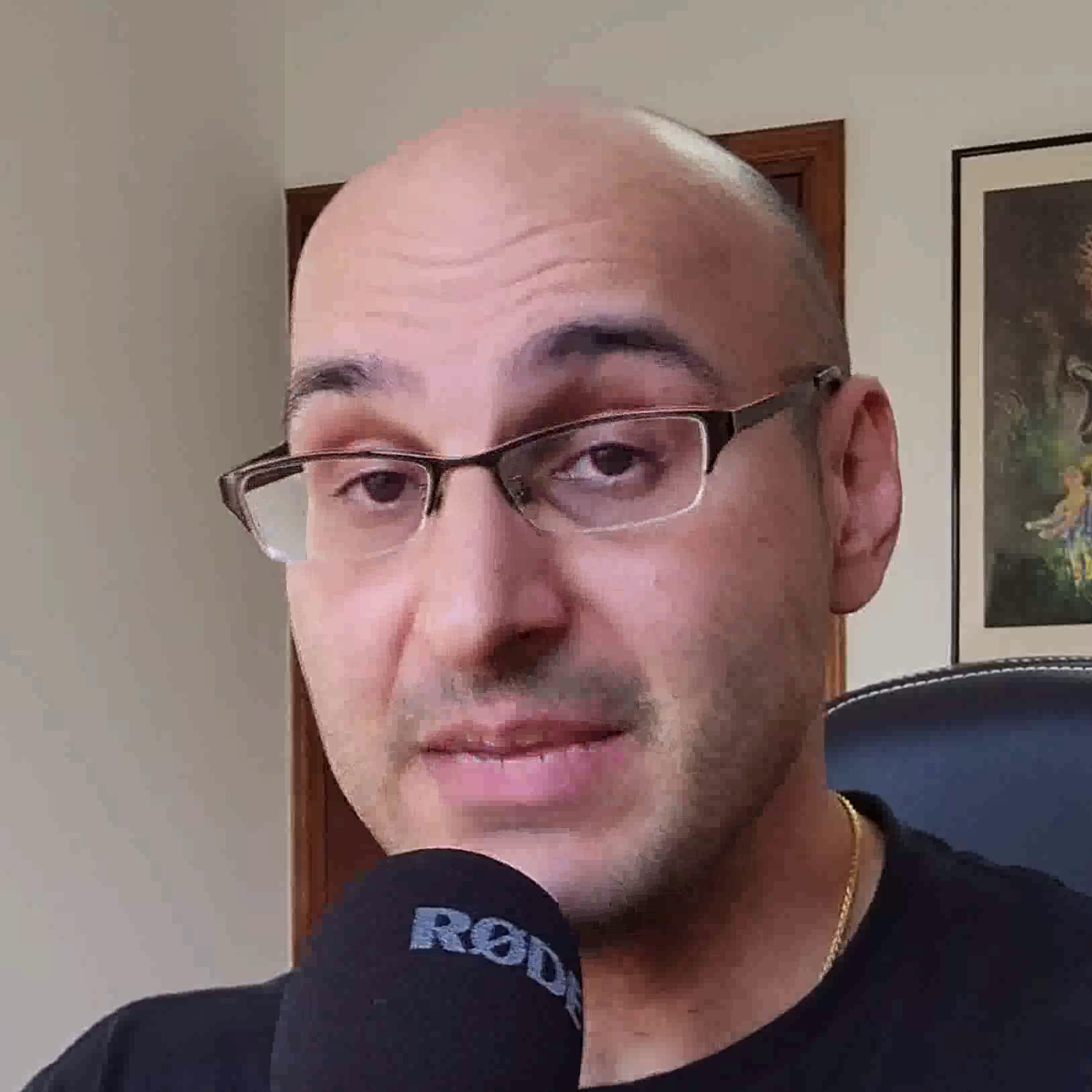} \\
        
        \raisebox{0.235in}{\rotatebox[origin=t]{90}{T2V-Zero}} &
        \includegraphics[clip,width=16mm]{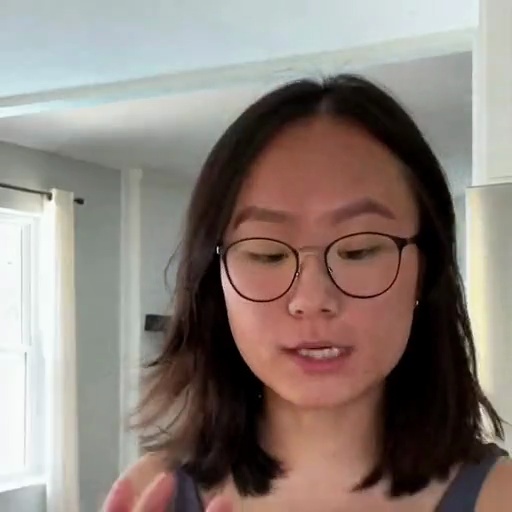} &
        \includegraphics[clip,width=16mm]{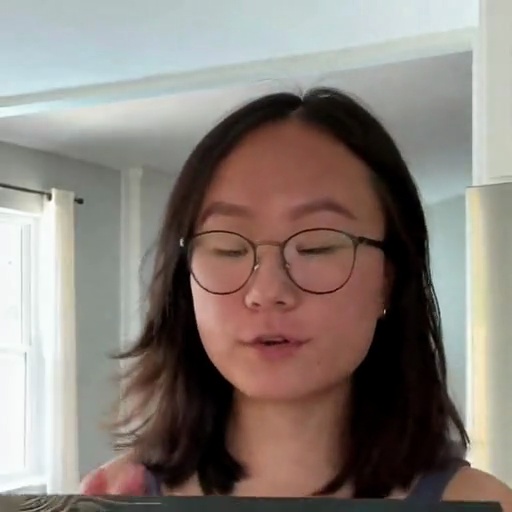} &
        \includegraphics[clip,width=16mm]{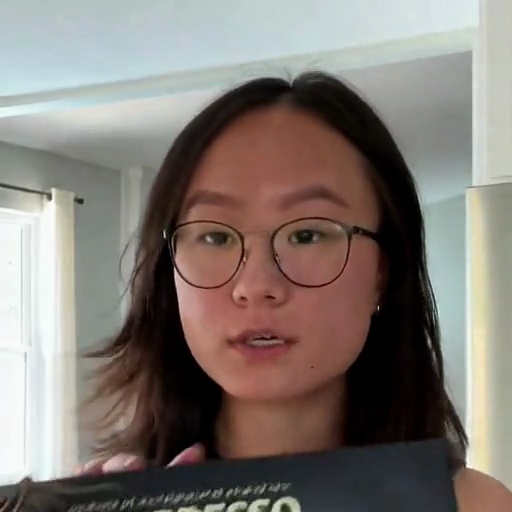} &
        \includegraphics[clip,width=16mm]{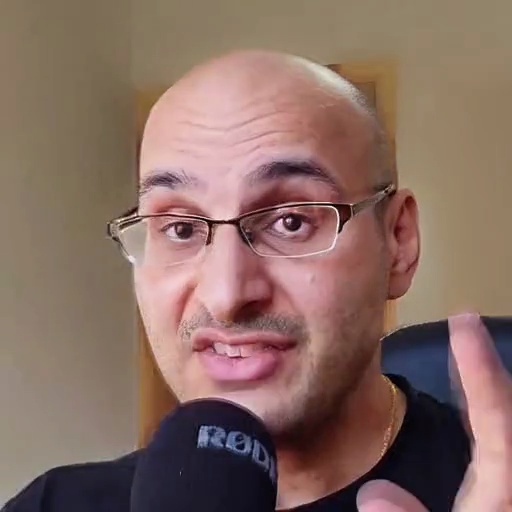} &
        \includegraphics[clip,width=16mm]{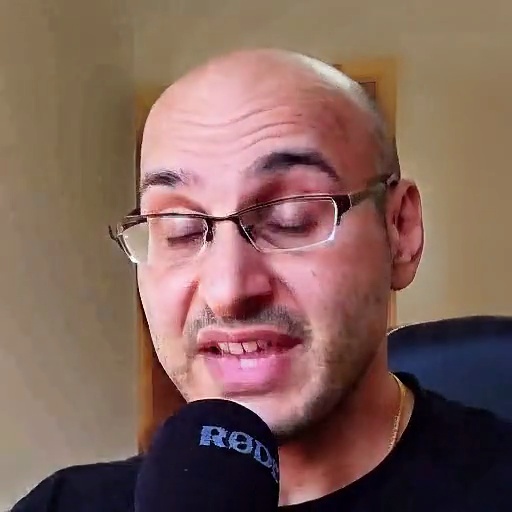} &
        \includegraphics[clip,width=16mm]{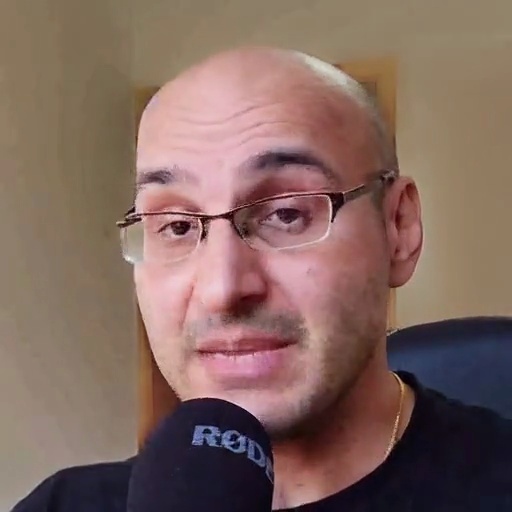} \\
        
        \raisebox{0.235in}{\rotatebox[origin=t]{90}{TokenFlow}} &
        \includegraphics[clip,width=16mm]{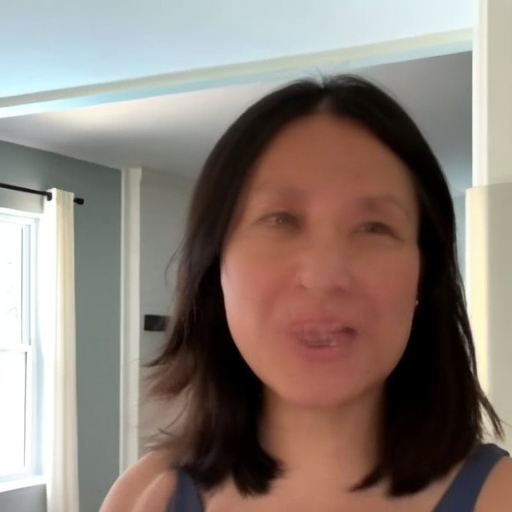} &
        \includegraphics[clip,width=16mm]{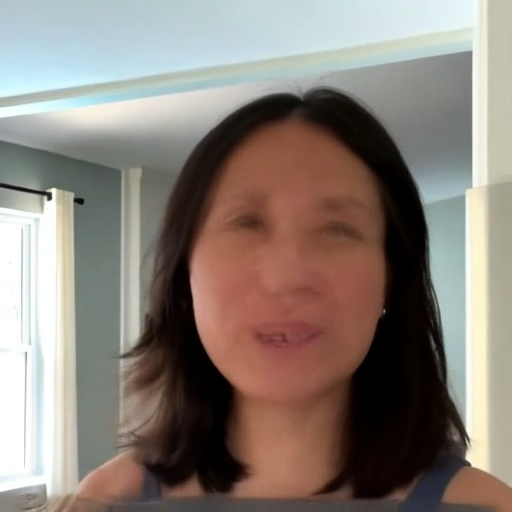} &
        \includegraphics[clip,width=16mm]{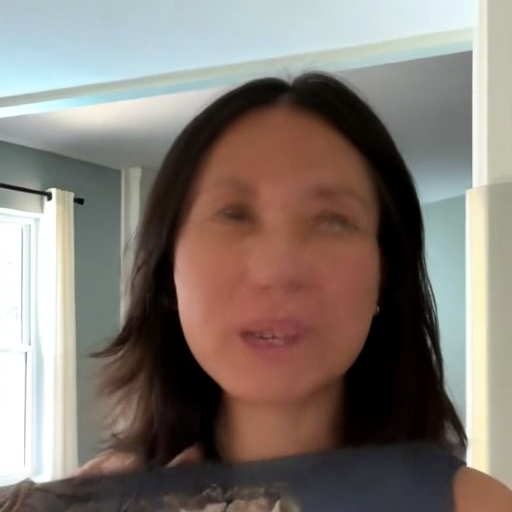} &
        \includegraphics[clip,width=16mm]{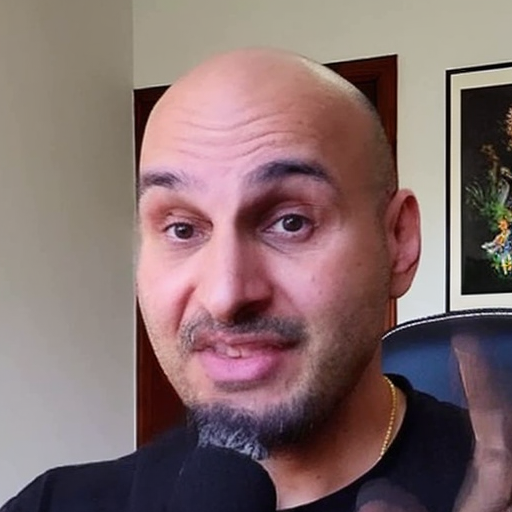} &
        \includegraphics[clip,width=16mm]{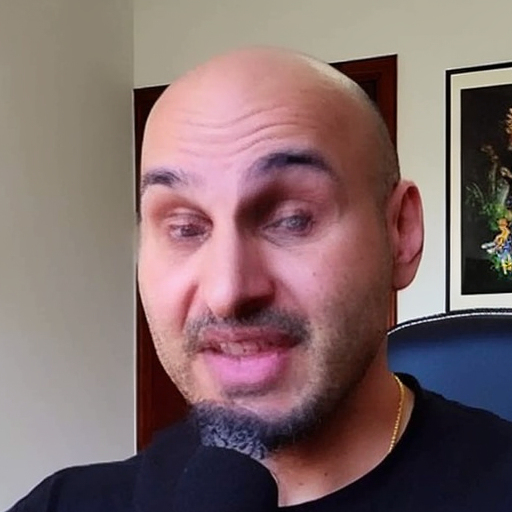} &
        \includegraphics[clip,width=16mm]{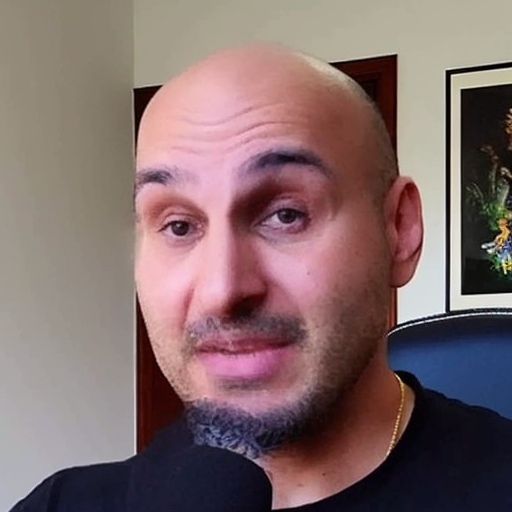} \\
        
        \raisebox{0.235in}{\rotatebox[origin=t]{90}{RAVE}} &
        \includegraphics[clip,width=16mm]{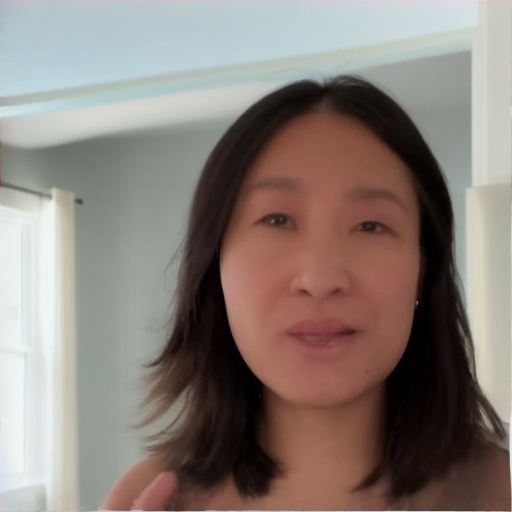} &
        \includegraphics[clip,width=16mm]{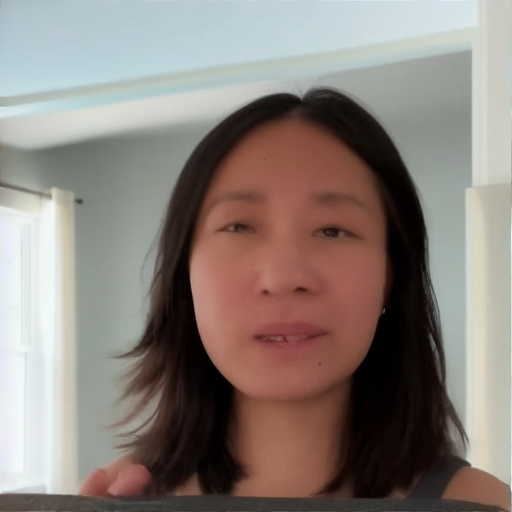} &
        \includegraphics[clip,width=16mm]{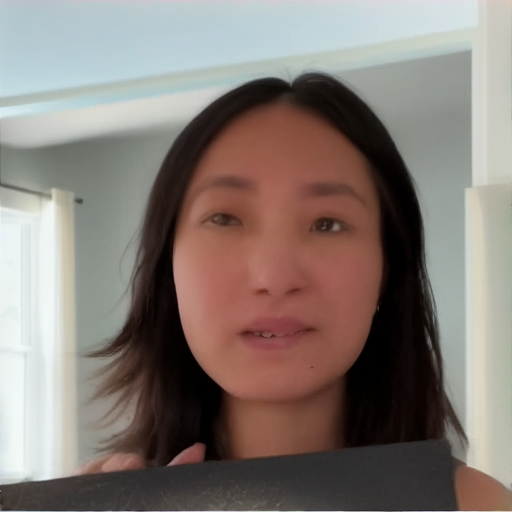} &
        \includegraphics[clip,width=16mm]{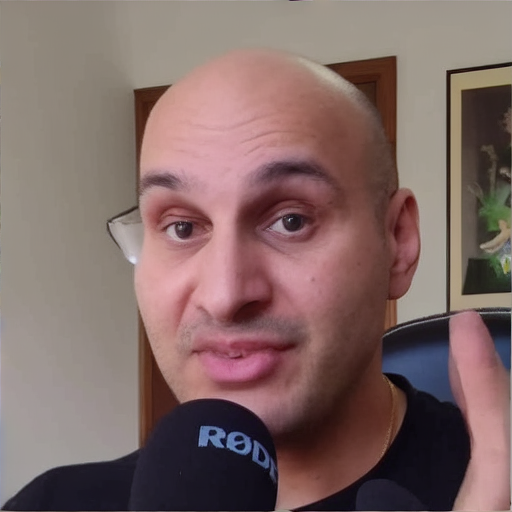} &
        \includegraphics[clip,width=16mm]{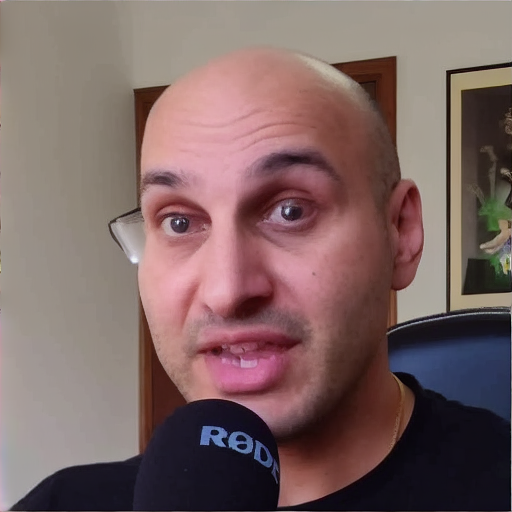} &
        \includegraphics[clip,width=16mm]{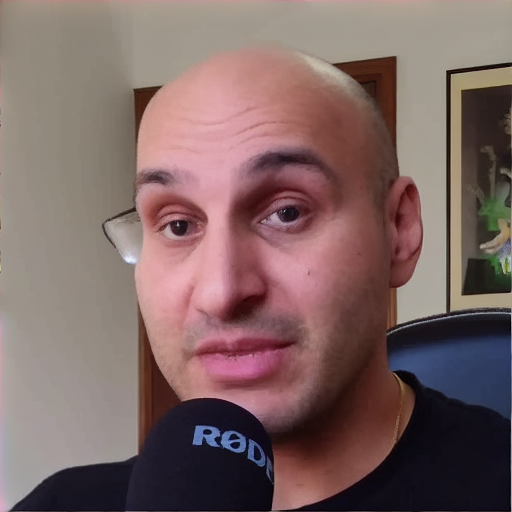} \\
        
        \raisebox{0.235in}{\rotatebox[origin=t]{90}{CN inpaint}} &
        \includegraphics[clip,width=16mm]{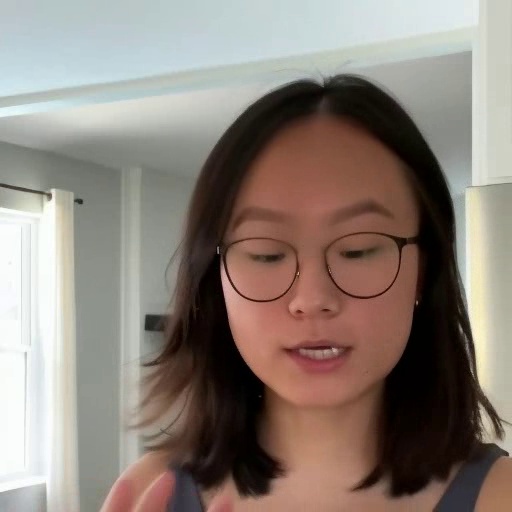} &
        \includegraphics[clip,width=16mm]{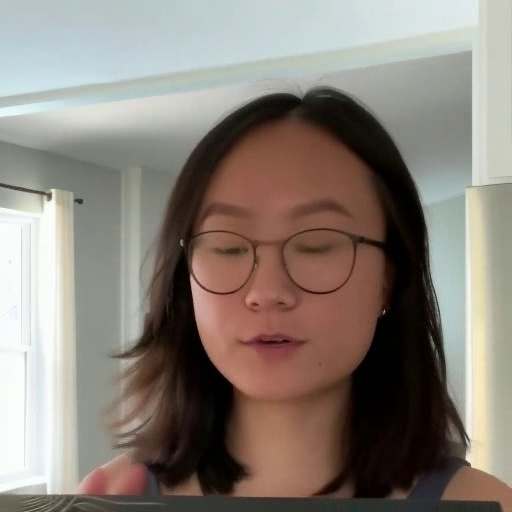} &
        \includegraphics[clip,width=16mm]{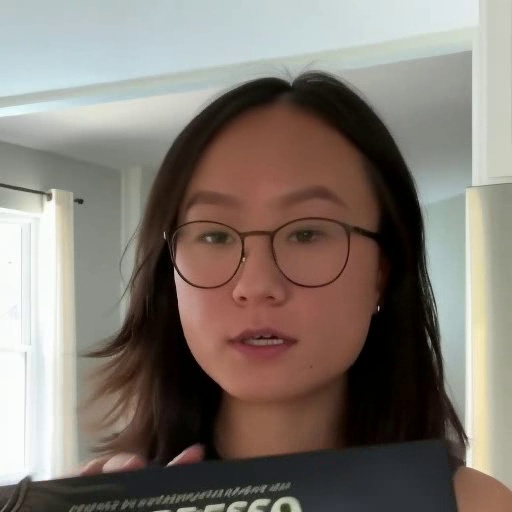} &
        \includegraphics[clip,width=16mm]{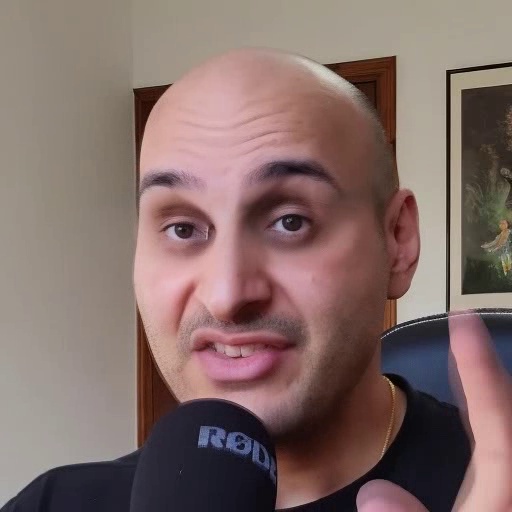} &
        \includegraphics[clip,width=16mm]{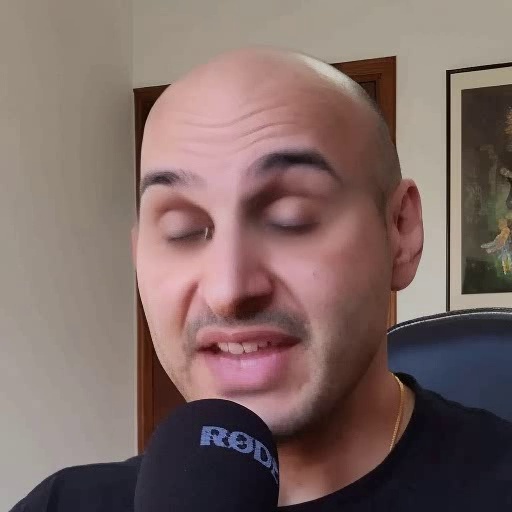} &
        \includegraphics[clip,width=16mm]{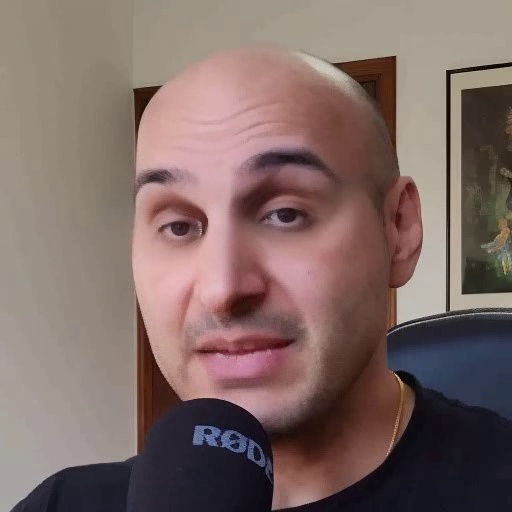} \\

        \raisebox{0.235in}{\rotatebox[origin=t]{90}{FGT}} &
        \includegraphics[clip,width=16mm]{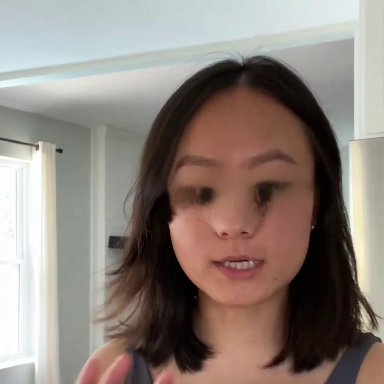} &
        \includegraphics[clip,width=16mm]{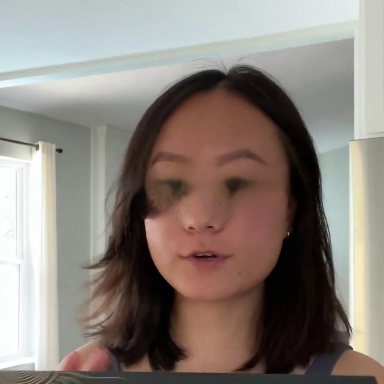} &
        \includegraphics[clip,width=16mm]{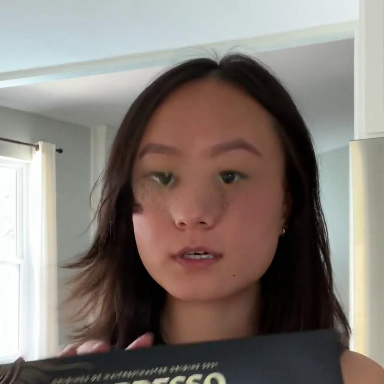} &
        \includegraphics[clip,width=16mm]{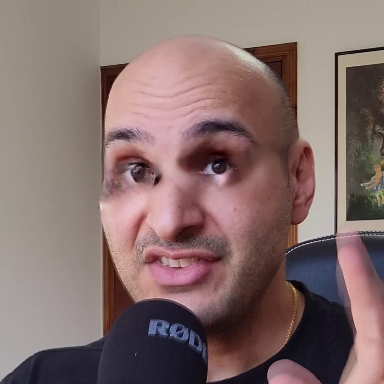} &
        \includegraphics[clip,width=16mm]{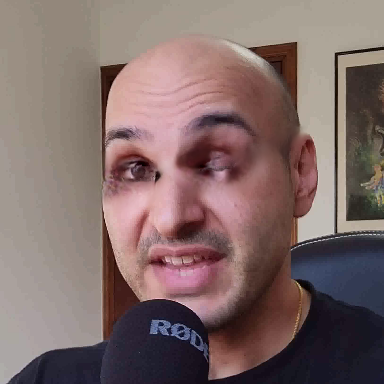} &
        \includegraphics[clip,width=16mm]{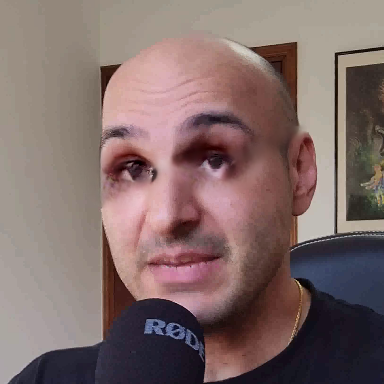} \\
        
       \raisebox{0.235in}{\rotatebox[origin=t]{90}{ProPainter}} &
       \includegraphics[clip,width=16mm]{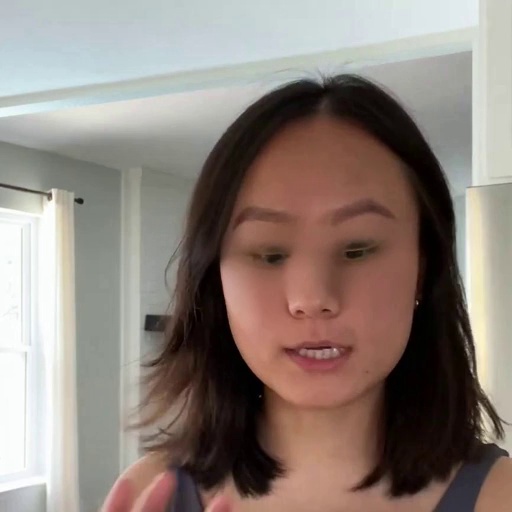} &
        \includegraphics[clip,width=16mm]{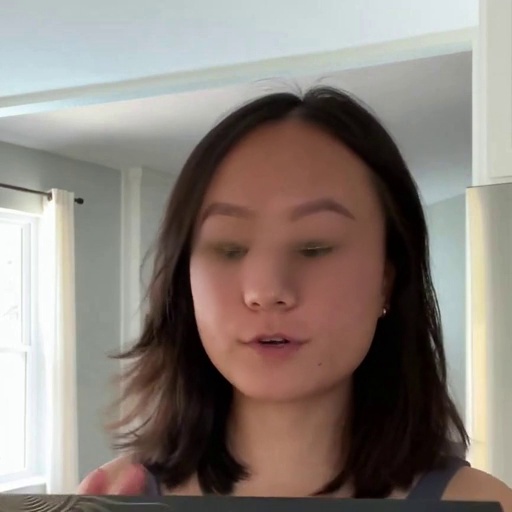} &
        \includegraphics[clip,width=16mm]{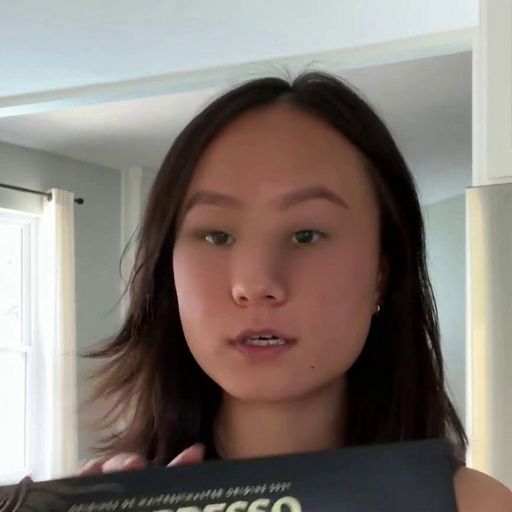} &
        \includegraphics[clip,width=16mm]{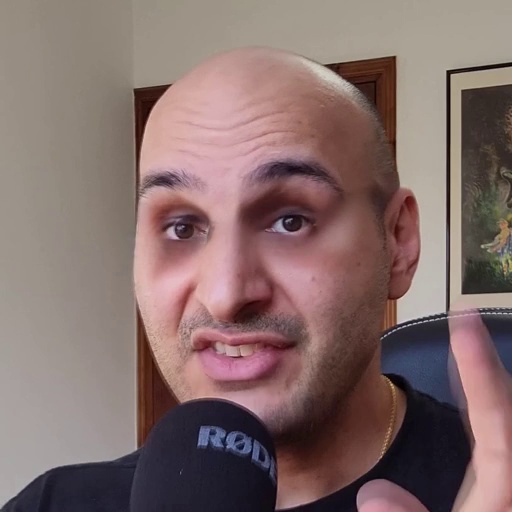} &
        \includegraphics[clip,width=16mm]{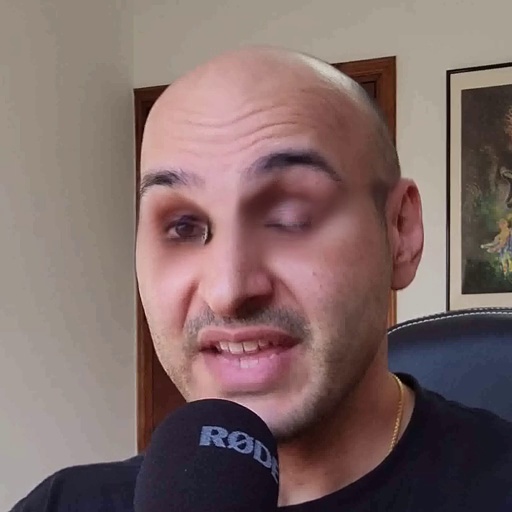} &
        \includegraphics[clip,width=16mm]{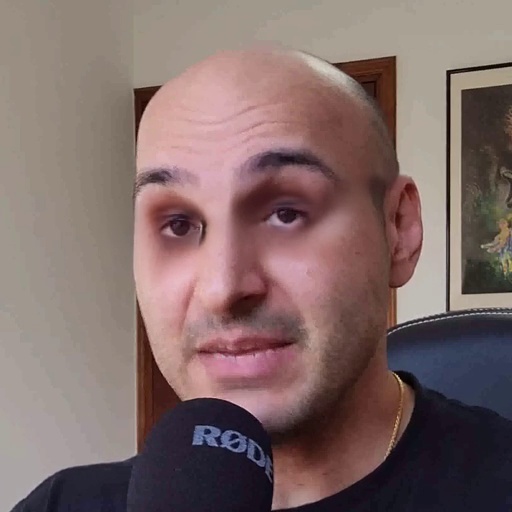} \\
        
       \raisebox{0.235in}{\rotatebox[origin=t]{90}{\textbf{Ours}}} &
        \includegraphics[clip,width=16mm]{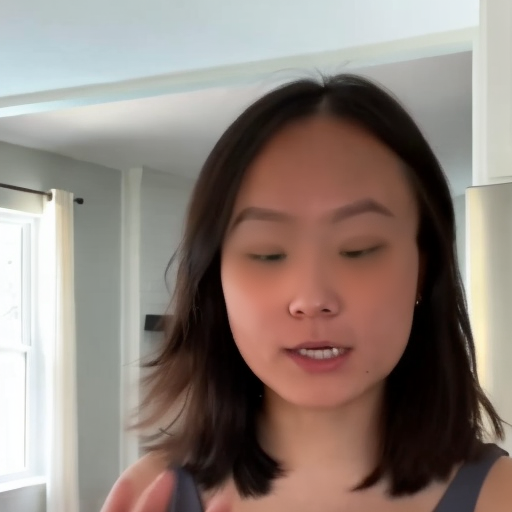} &
        \includegraphics[clip,width=16mm]{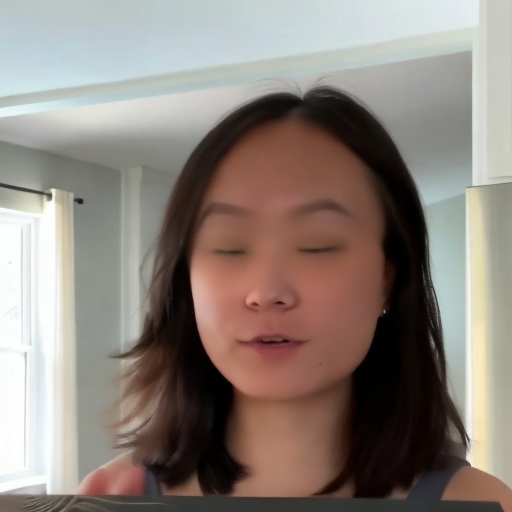} &
        \includegraphics[clip,width=16mm]{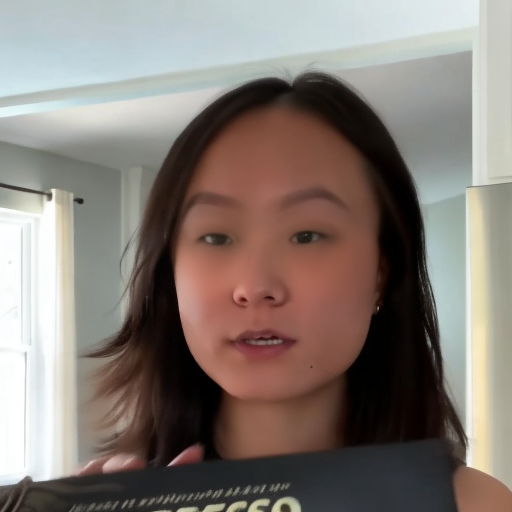} &
        \includegraphics[clip,width=16mm]{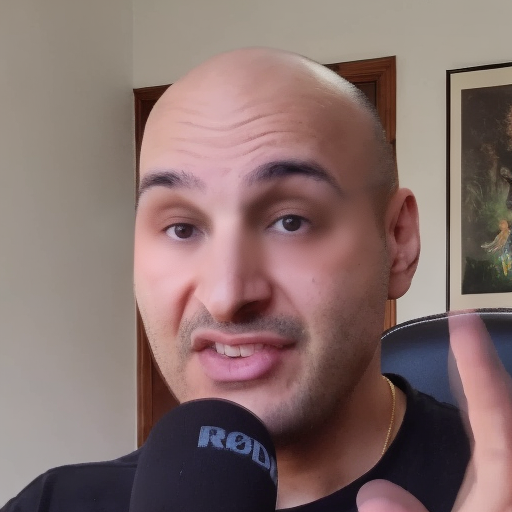} &
        \includegraphics[clip,width=16mm]{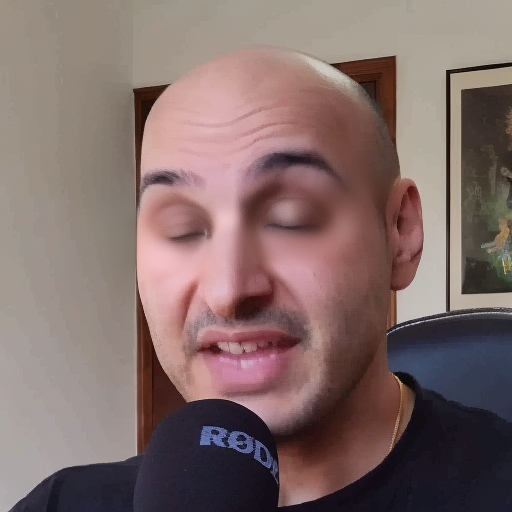} &
        \includegraphics[clip,width=16mm]{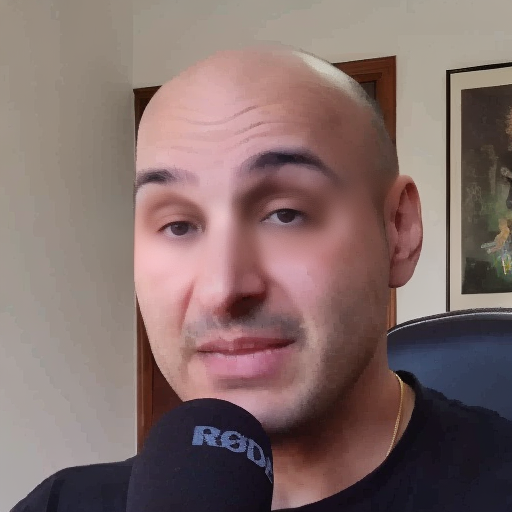} \\
    \end{tabular}
    }
    \caption{\textbf{Visual comparisons:} We compare our results to different video editing and inpainting methods. Other methods often struggle with glasses-removal, and even when they do remove the glasses, they tend to leave glasses remnants (e.g. RAVE right example), generate artifacts (e.g. FGT, ProPainter examples, TokenFlow left example), do not preserve the identity of the person (e.g. RAVE Left example), or their eyelids position (e.g. RAVE both examples).
    }
    \label{fig:results_comp}
       \Description[]{}  % use this line to please the compiler
\end{figure*}
\begin{figure*}[htpb]
    \centering
    \setlength{\tabcolsep}{0.1pt}

    \resizebox{.86\textwidth}{!}{
    \begin{tabular}{c@{\hskip 0.5em} c c c@{\hskip 0.5em} c c c}

        \raisebox{0.235in}{\rotatebox[origin=t]{90}{Input}} &
        \includegraphics[clip,width=16mm]{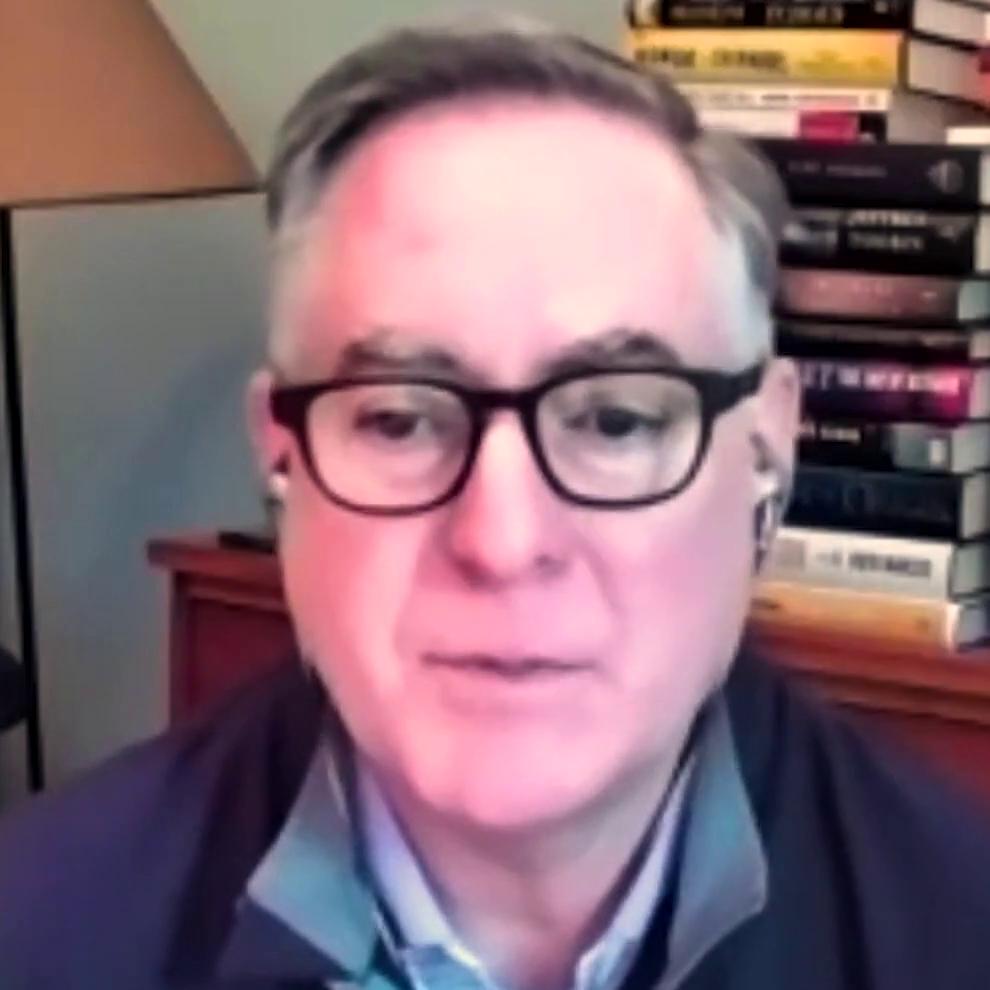} &
        \includegraphics[clip,width=16mm]{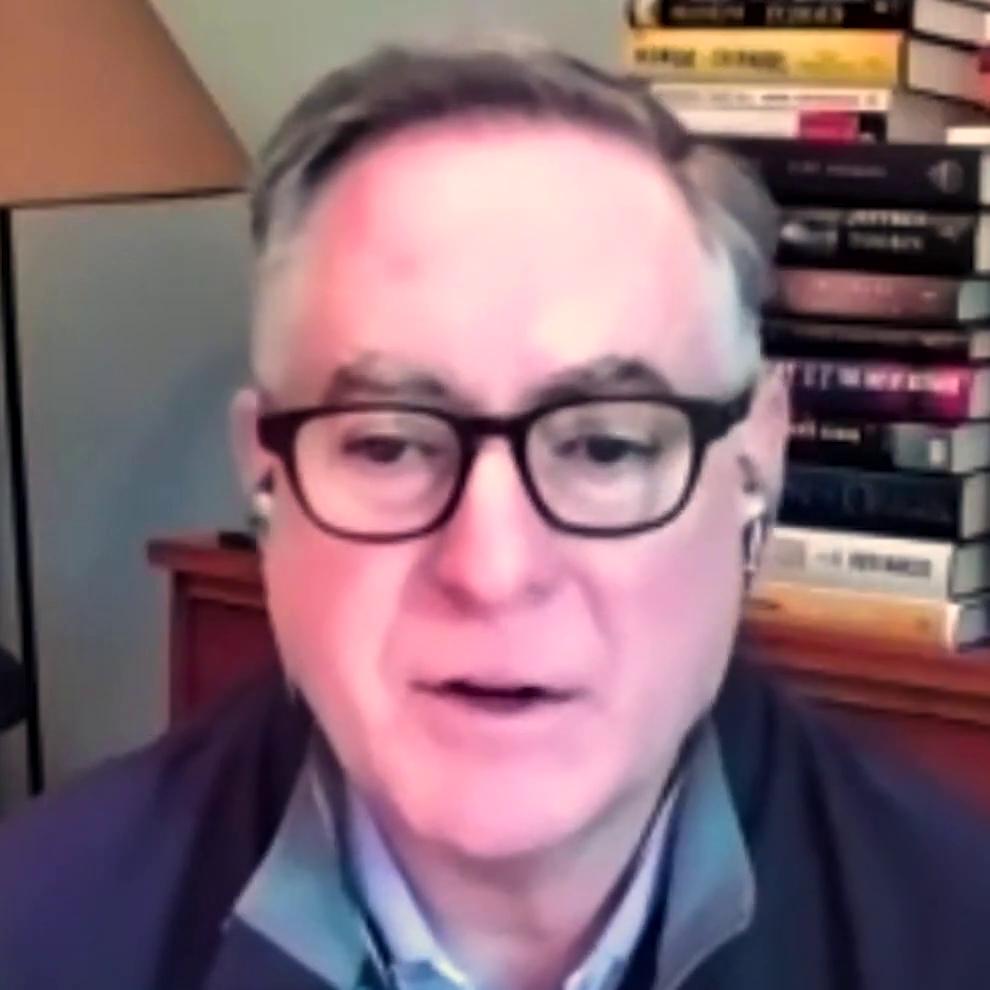} &
        \includegraphics[clip,width=16mm]{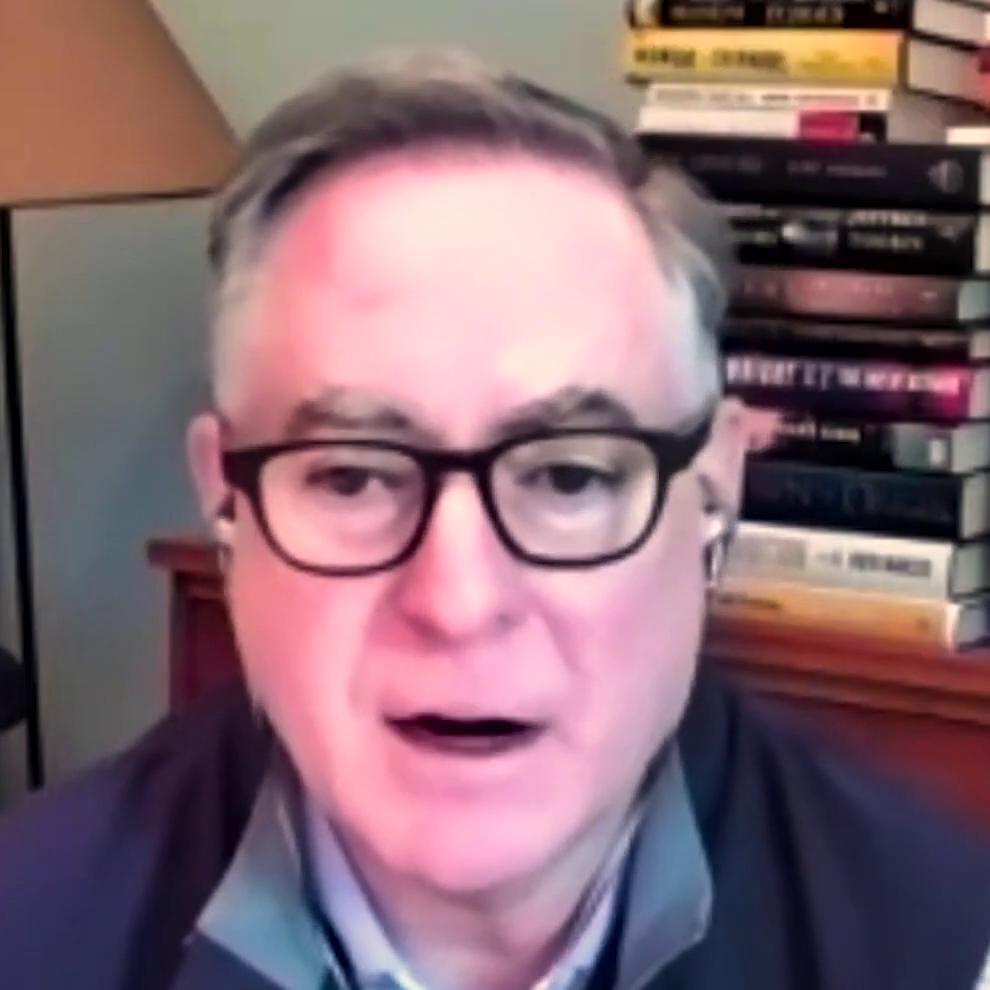} &
        \includegraphics[clip,width=16mm]{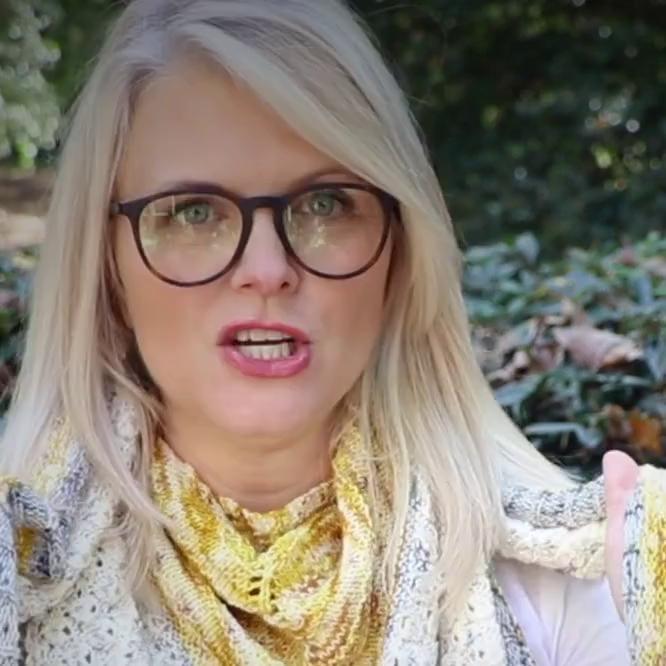} &
        \includegraphics[clip,width=16mm]{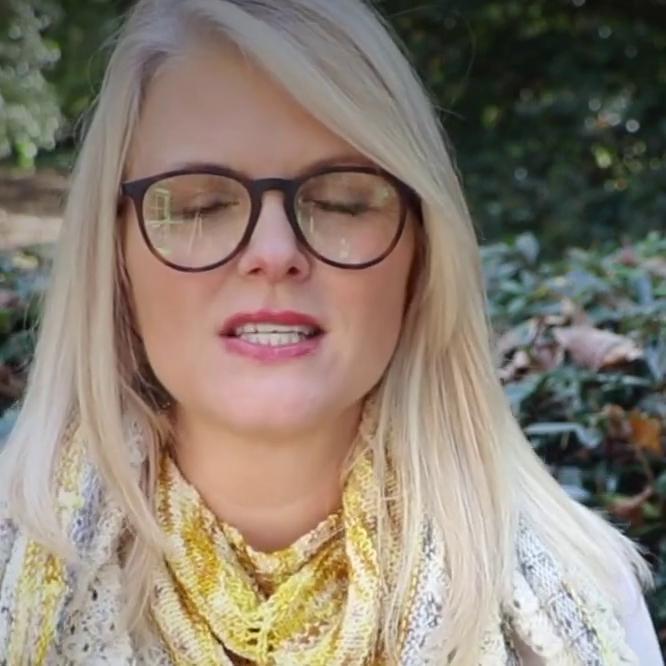} &
        \includegraphics[clip,width=16mm]{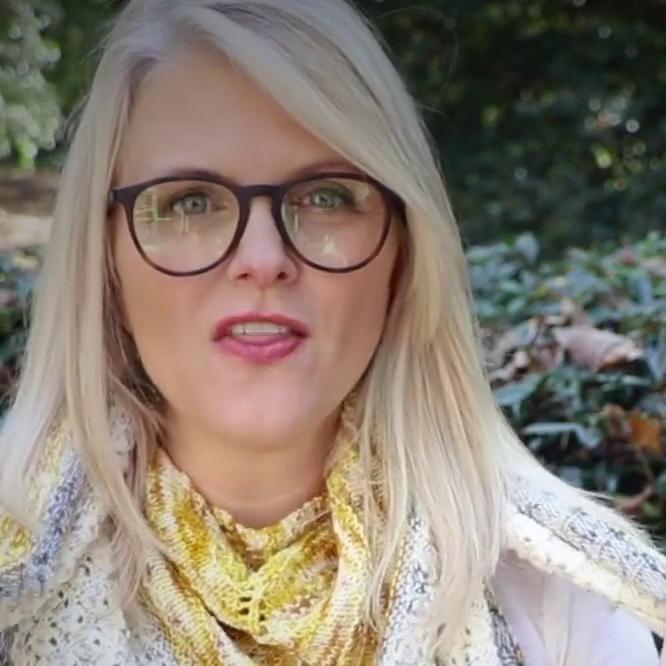} \\
        
        \raisebox{0.235in}{\rotatebox[origin=t]{90}{T2V-Zero}} &
        \includegraphics[clip,width=16mm]{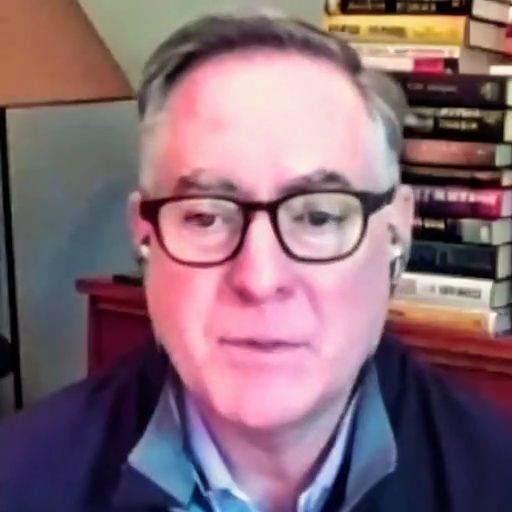} &
        \includegraphics[clip,width=16mm]{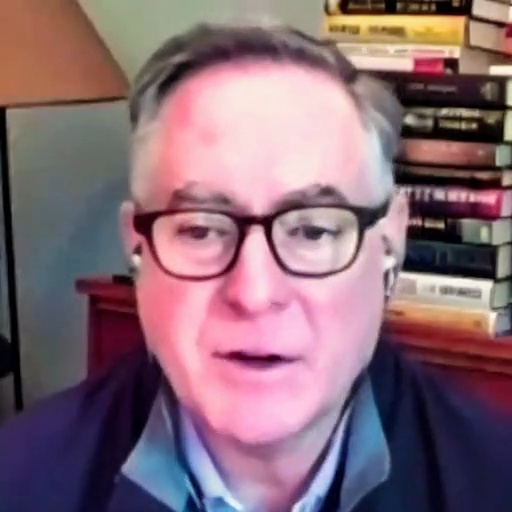} &
        \includegraphics[clip,width=16mm]{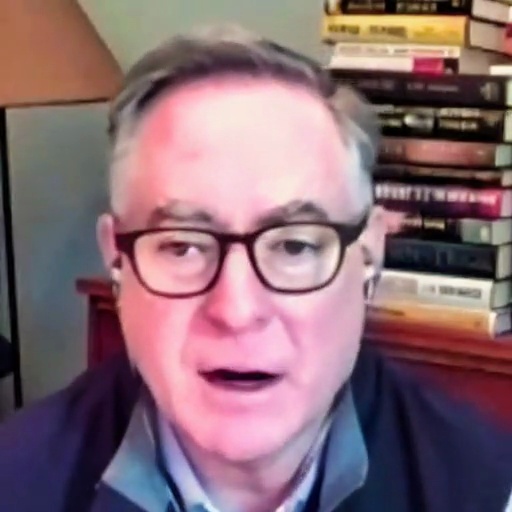} &
        \includegraphics[clip,width=16mm]{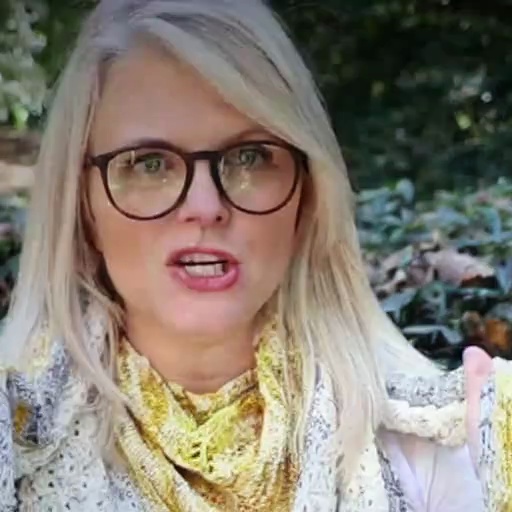} &
        \includegraphics[clip,width=16mm]{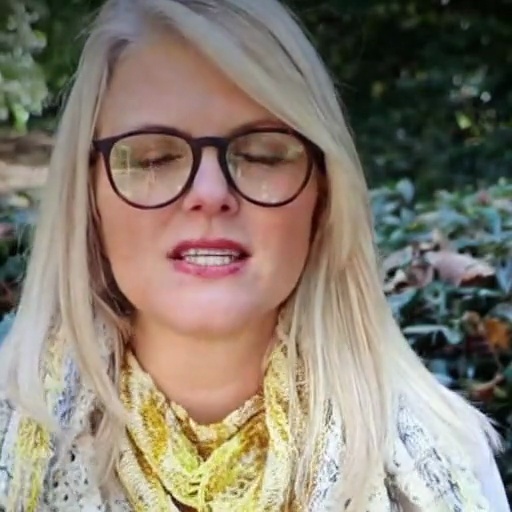} &
        \includegraphics[clip,width=16mm]{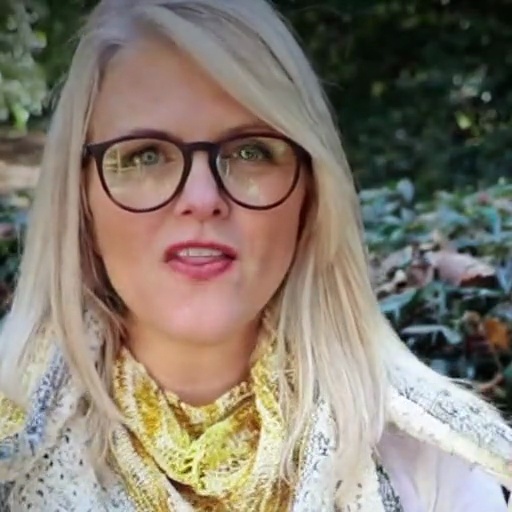} \\
        
        \raisebox{0.235in}{\rotatebox[origin=t]{90}{TokenFlow}} &
       \includegraphics[clip,width=16mm]{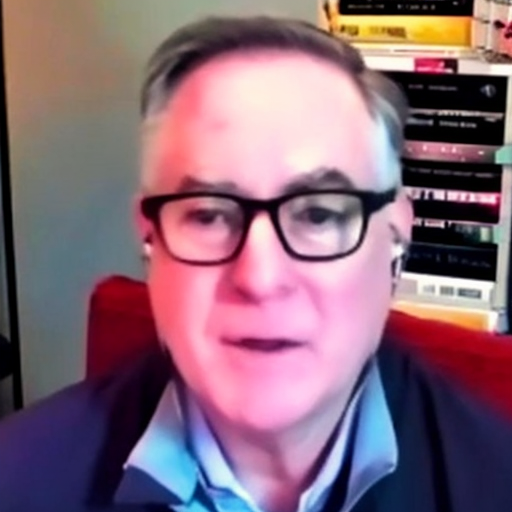} &
        \includegraphics[clip,width=16mm]{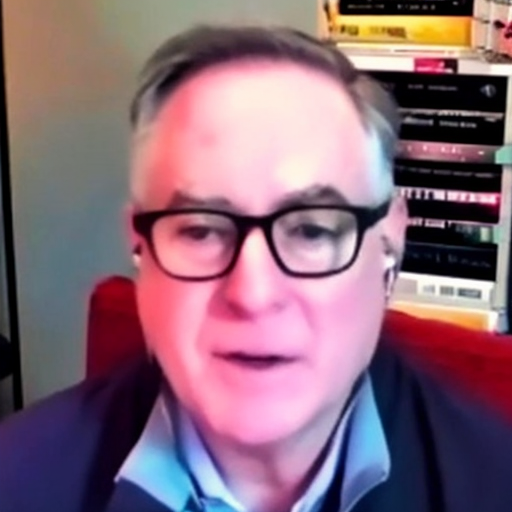} &
        \includegraphics[clip,width=16mm]{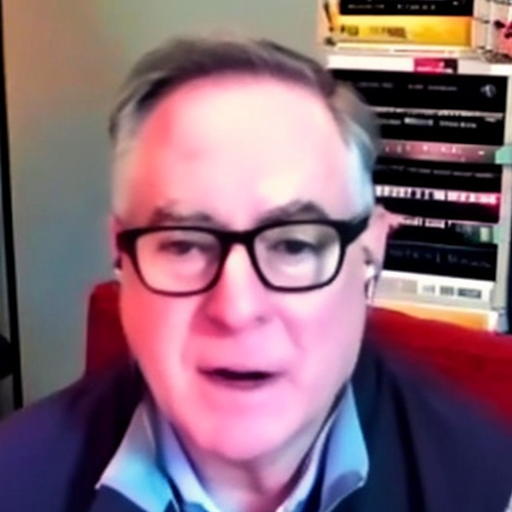} &
        \includegraphics[clip,width=16mm]{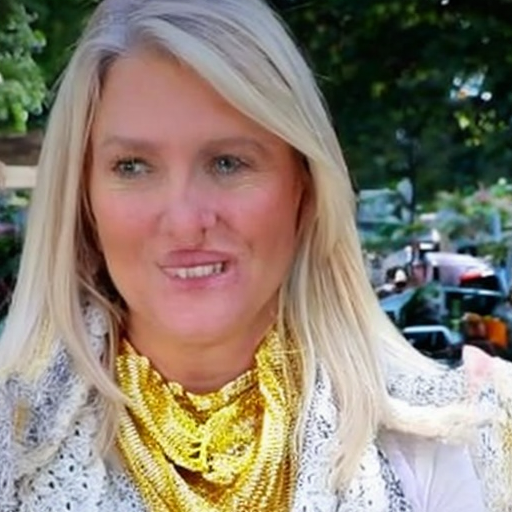} &
        \includegraphics[clip,width=16mm]{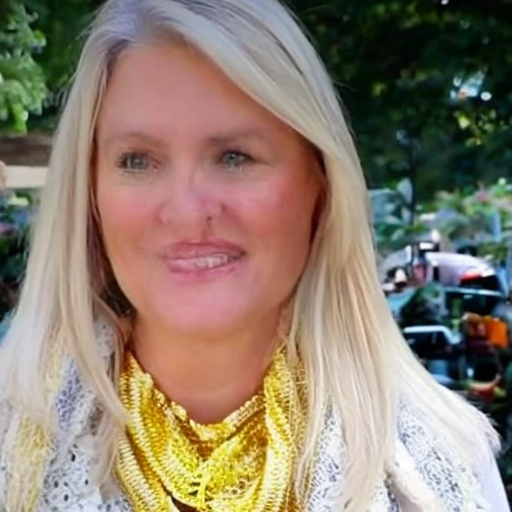} &
        \includegraphics[clip,width=16mm]{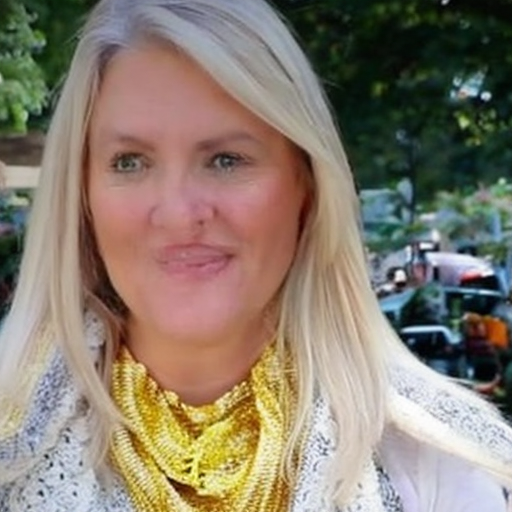} \\
        
        \raisebox{0.235in}{\rotatebox[origin=t]{90}{RAVE}} &
       \includegraphics[clip,width=16mm]{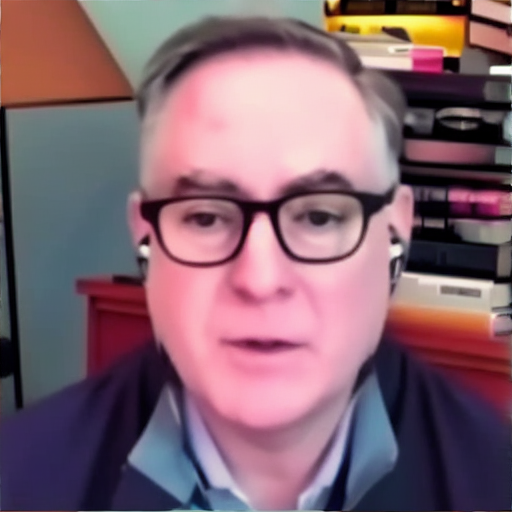} &
        \includegraphics[clip,width=16mm]{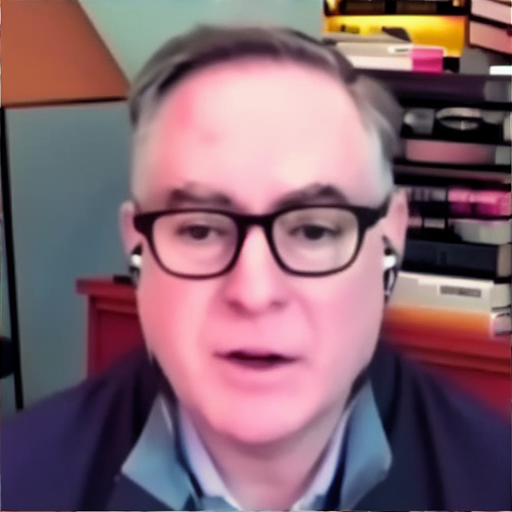} &
        \includegraphics[clip,width=16mm]{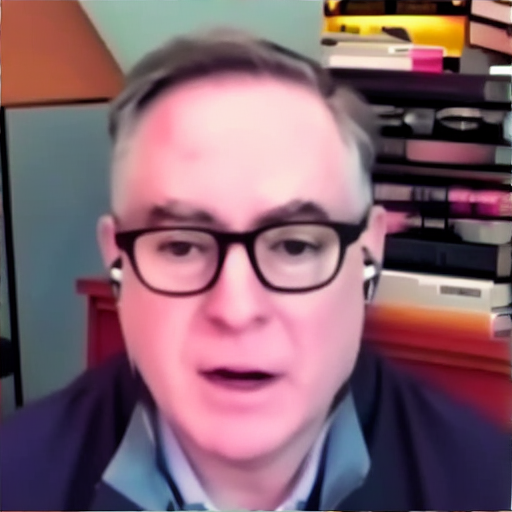} &
        \includegraphics[clip,width=16mm]{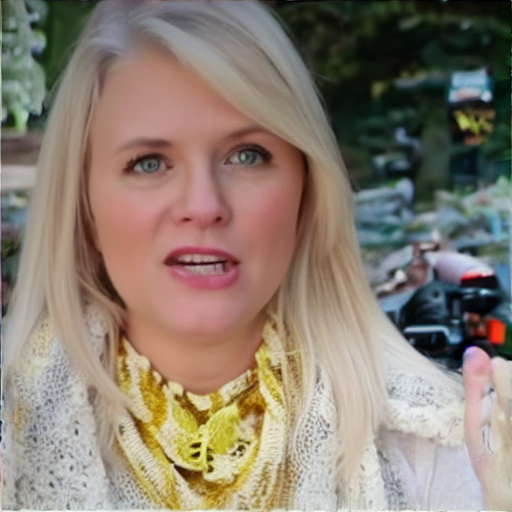} &
        \includegraphics[clip,width=16mm]{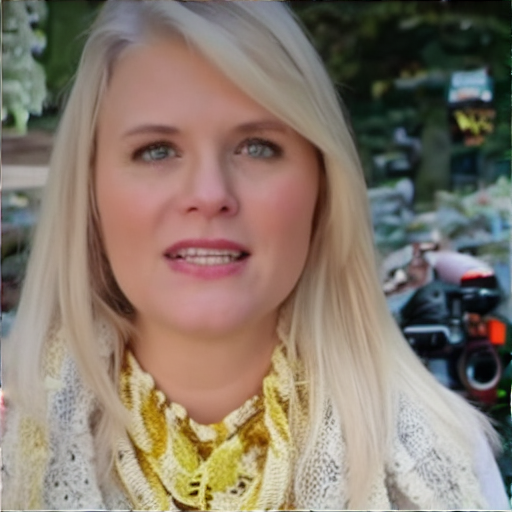} &
        \includegraphics[clip,width=16mm]{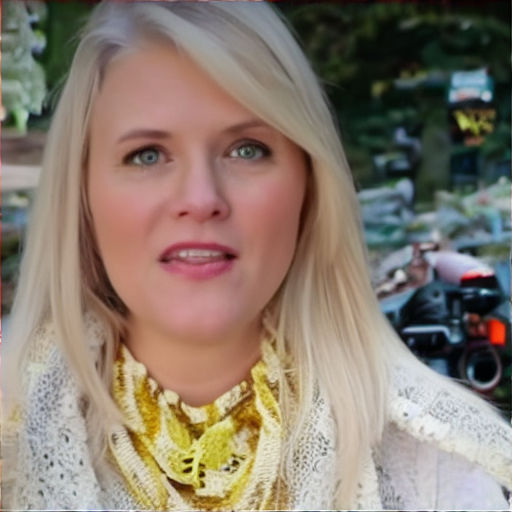} \\
        
        \raisebox{0.235in}{\rotatebox[origin=t]{90}{CN inpaint}} &
       \includegraphics[clip,width=16mm]{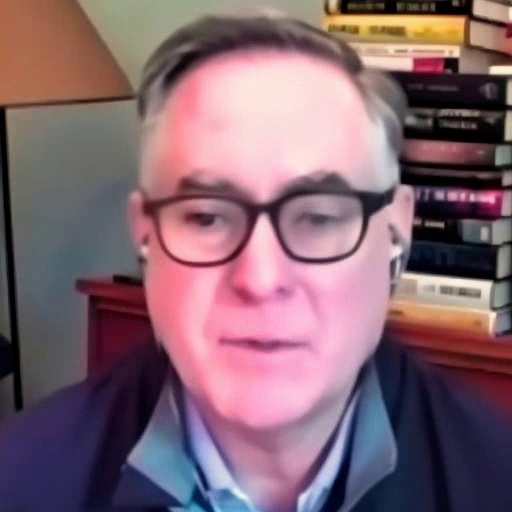} &
        \includegraphics[clip,width=16mm]{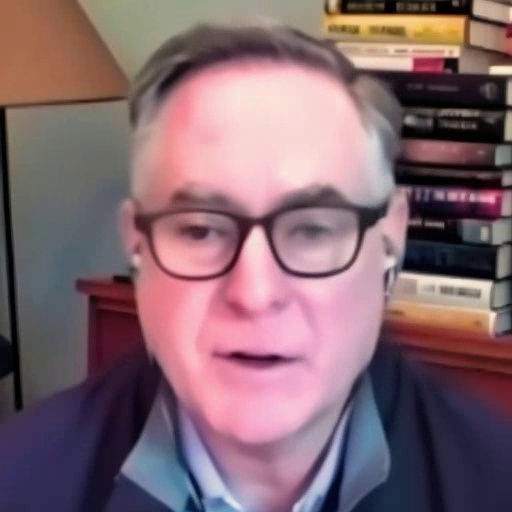} &
        \includegraphics[clip,width=16mm]{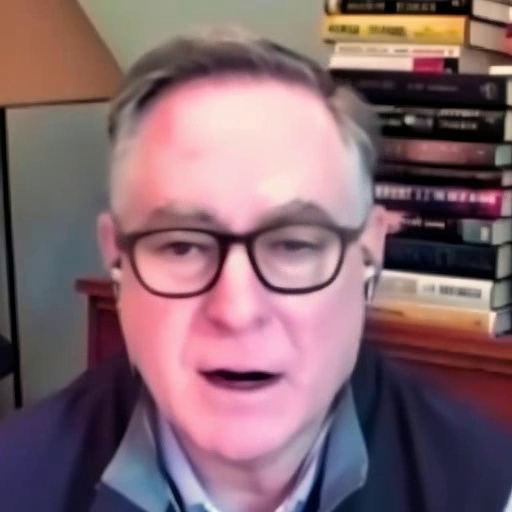} &
       \includegraphics[clip,width=16mm]{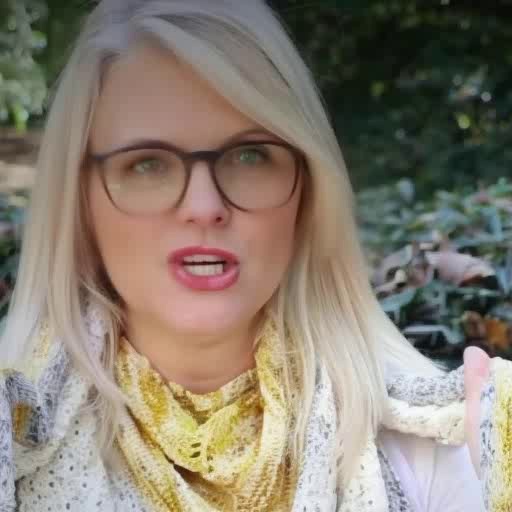} &
        \includegraphics[clip,width=16mm]{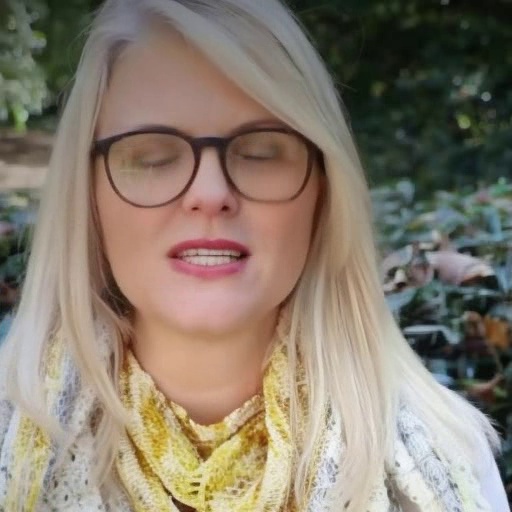} &
        \includegraphics[clip,width=16mm]{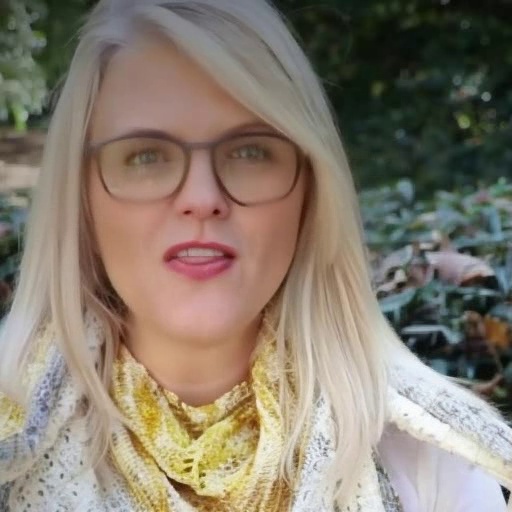} \\
        
       \raisebox{0.235in}{\rotatebox[origin=t]{90}{FGT}} &
       \includegraphics[clip,width=16mm]{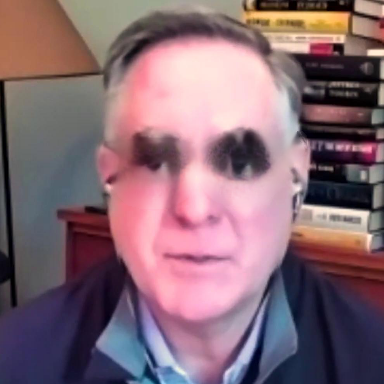} &
        \includegraphics[clip,width=16mm]{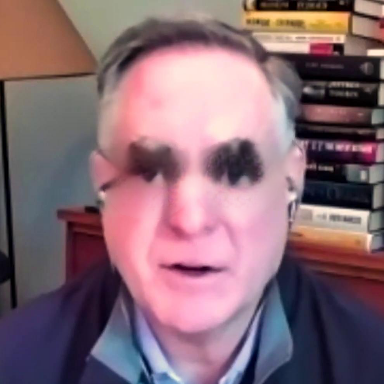} &
        \includegraphics[clip,width=16mm]{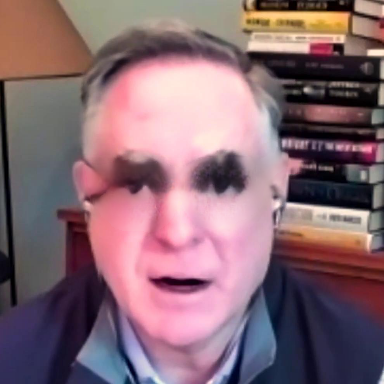} &
        \includegraphics[clip,width=16mm]{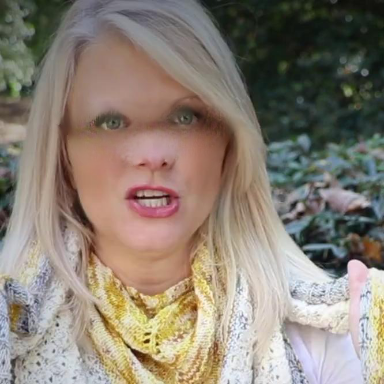} &
        \includegraphics[clip,width=16mm]{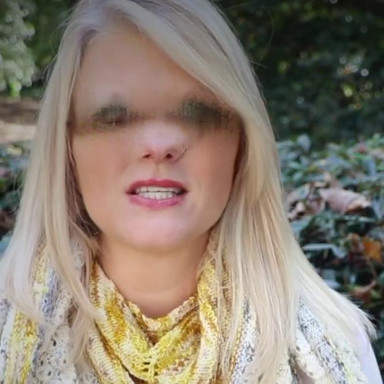} &
        \includegraphics[clip,width=16mm]{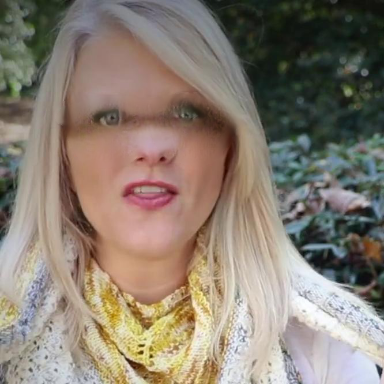} \\
        
       \raisebox{0.235in}{\rotatebox[origin=t]{90}{ProPainter}} &
       \includegraphics[clip,width=16mm]{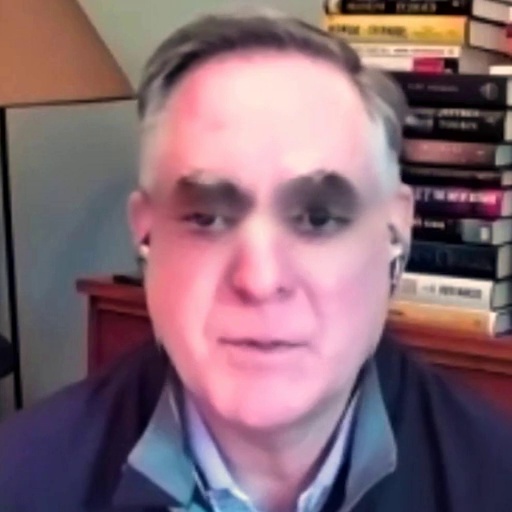} &
        \includegraphics[clip,width=16mm]{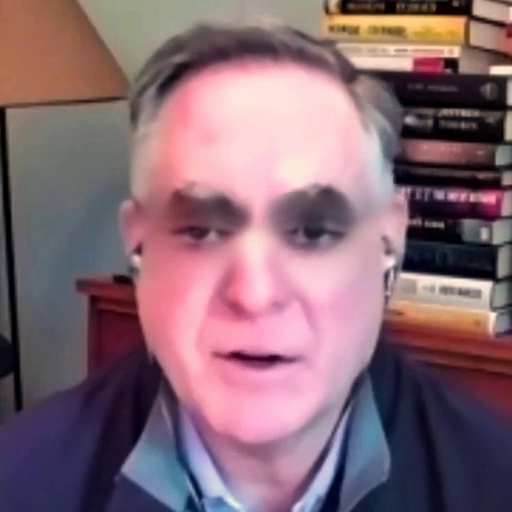} &
        \includegraphics[clip,width=16mm]{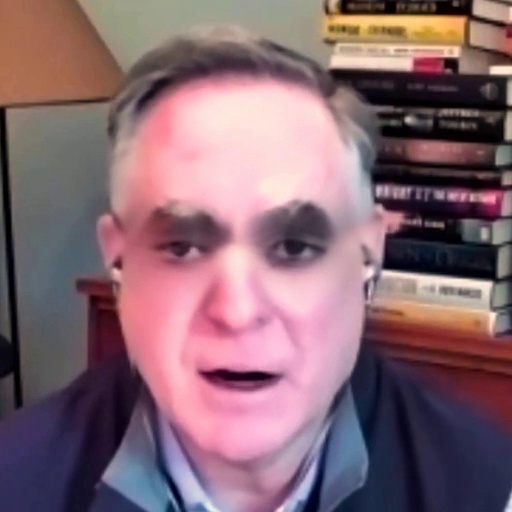} &
        \includegraphics[clip,width=16mm]{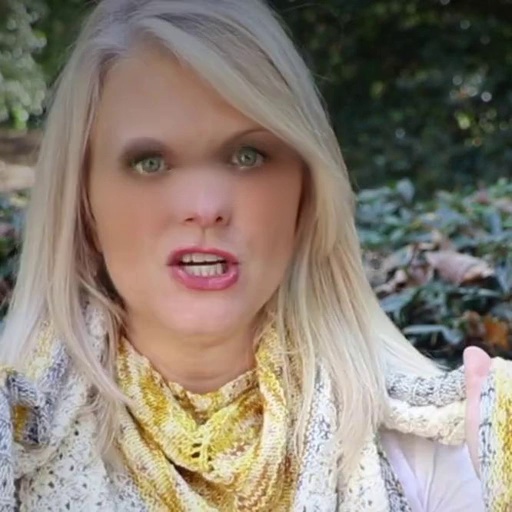} &
        \includegraphics[clip,width=16mm]{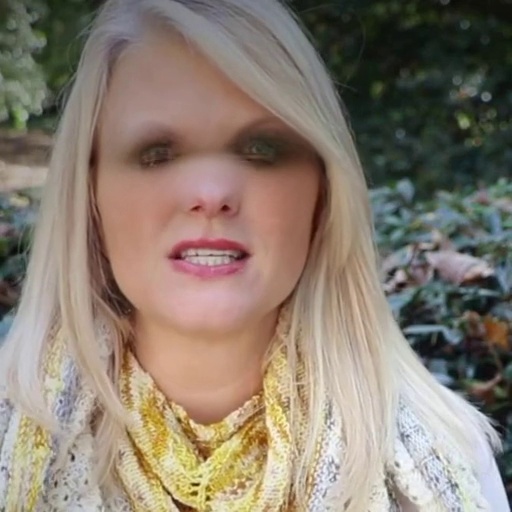} &
        \includegraphics[clip,width=16mm]{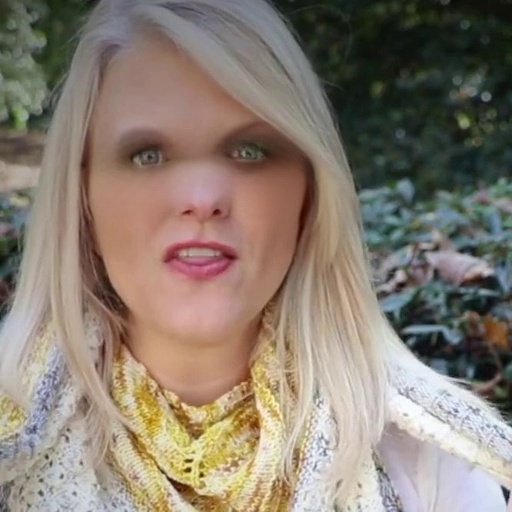}  \\

        \raisebox{0.235in}{\rotatebox[origin=t]{90}{\textbf{Ours}}} &
       \includegraphics[clip,width=16mm]{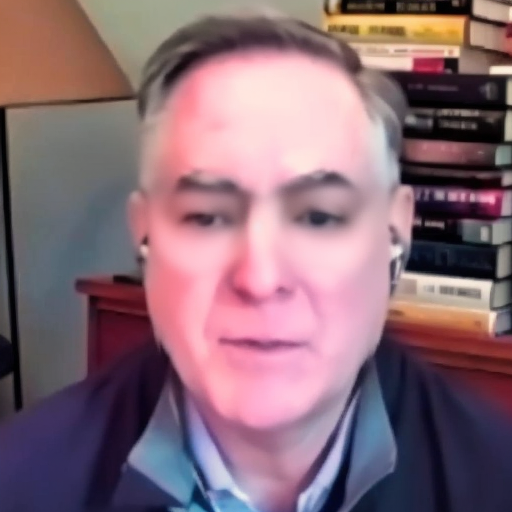} &
        \includegraphics[clip,width=16mm]{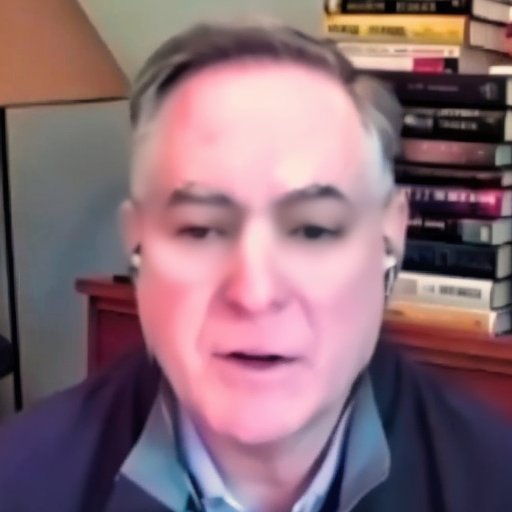} &
        \includegraphics[clip,width=16mm]{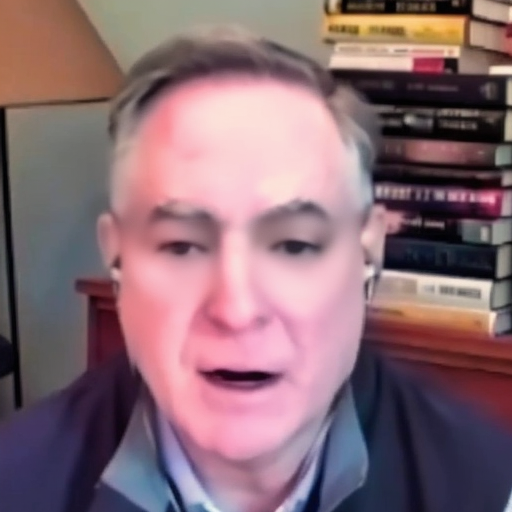} &
       \includegraphics[clip,width=16mm]{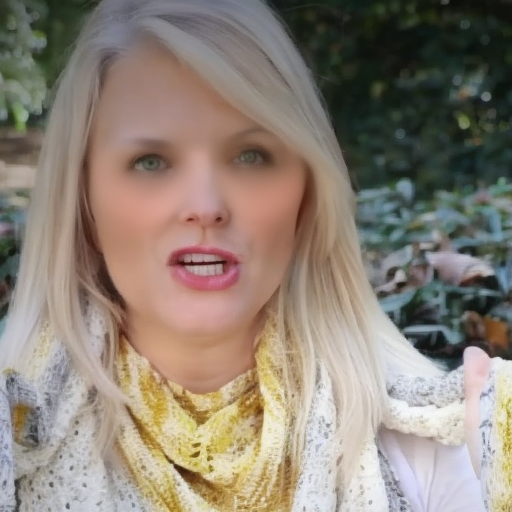} &
        \includegraphics[clip,width=16mm]{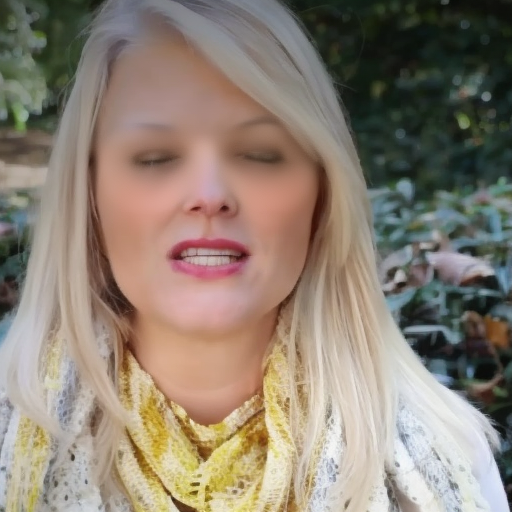} &
        \includegraphics[clip,width=16mm]{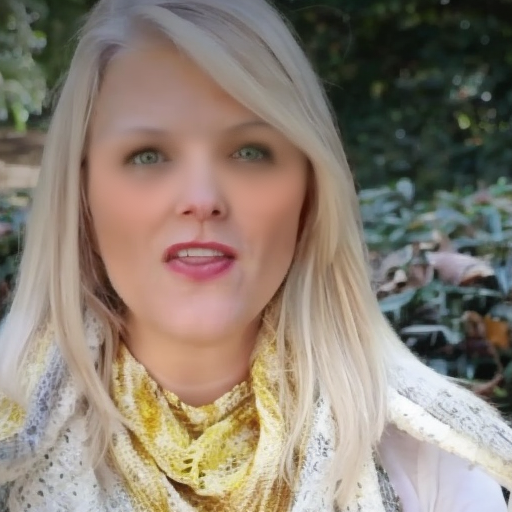} \\
    \end{tabular}
    }
    \caption{\textbf{Visual comparison:} We compare our results to different video editing and inpainting methods. Other methods often struggle with glasses-removal, and even when they do remove the glasses, they tend to either change the identity completely (e.g. TokenFlow right example), generate artifacts such as black areas around the eyes (e.g. FGT, ProPainter both examples), or do not preserve the eyelids position (e.g. TokenFlow, RAVE right example)}.
    \label{fig:results_comp2}
       \Description[]{}  % use this line to please the compiler
\end{figure*}

We evaluate our results both qualitatively and quantitatively, testing three aspects: \\
1. \textbf{Fidelity} to the required edit, i.e glasses-removal. \\
2. \textbf{Identity and content preservation} of the original video. We want to remove the glasses while leaving the rest of the video intact. \\
3. \textbf{Realism} of the result, by means of temporal consistency and realism per-frame.

As we perform local video editing, we compare our results with the results of the SOTA video editing methods: 
TokenFlow \cite{geyer2023tokenflow}, RAVE \cite{kara2023rave}, and Text2Video-Zero \cite{khachatryan2023text2video}. TokenFlow and Text2Video-Zero incorporate SDEdit \cite{meng2021sdedit} and instructPix2Pix \cite{brooks2023instructpix2pix}, respectively, to allow for local attribute editing, and RAVE uses CN ``depth-zoe'' \cite{zhang2023adding} for that purpose. 
We also compare our model to the SOTA video inpainting works ProPainter \cite{zhou2023propainter} and FGT \cite{zhang2022flow} 
as our work is similar to inpainting works in the sense that it tries to replace some part of the video. Additionally, we compare our results to the results of our video editing pipeline with CN inpaint \cite{zhang2023adding} instead of our trained model, to emphasize that existing image inpainting models do not perform well enough on this task, even when combined in our video editing pipeline.
Our quantitative and qualitative evaluations, including a comprehensive user study, demonstrate that our results are favored over all other methods across all tested aspects.

\subsection{Experimental details}
\label{subsec:eval_details}

We generate our data pairs as described in \cref{subsec:data} from the dataset CelebV-Text \cite{yu2023celebv}. We train our model over 1296 of those videos, and test over 144 unseen videos.

\subsection{Qualitative evaluation}
\label{subsec:qual_eval}
\begin{figure}[htpb]
    \centering
    \begin{tabular}{c}
    \includegraphics[width=\columnwidth]{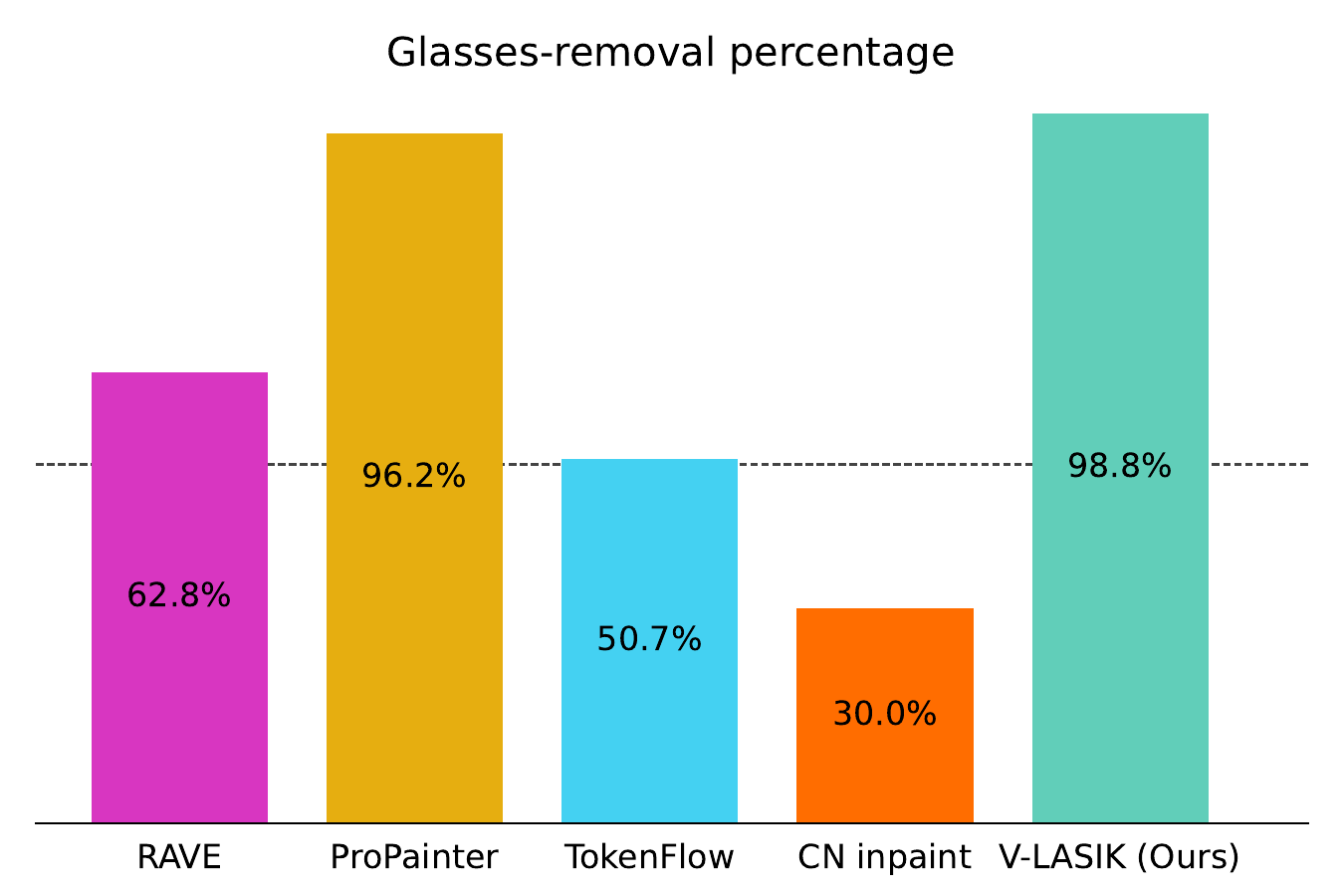} \\
    \\
\includegraphics[width=\columnwidth]{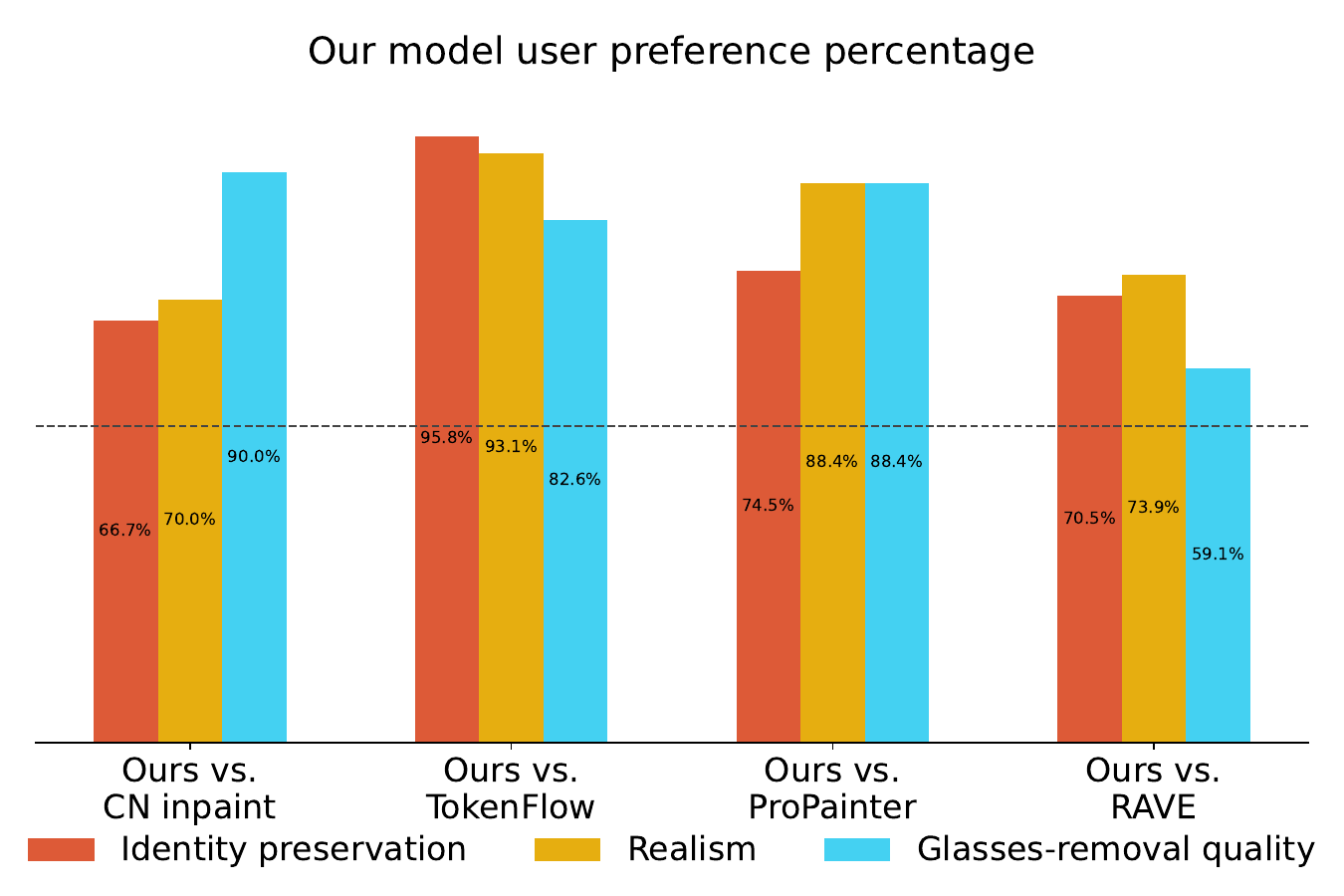} \\
    \end{tabular}
    \caption{\textbf{User study results:} First, we ask users which models remove the glasses from the input video (top). Then, when both models remove the glasses, we ask 
    which one better preserves the identity of the person, which one is more realistic, and which result has less remnants of glasses (bottom). Compared to all model examined, the users preferred our results through all measured aspects.}
    \label{fig:user_study}
       \Description[]{}  % use this line to please the compiler
\end{figure}

Visual results of our method are presented in \cref{fig:results_comp,fig:results_comp2} and in the supplementary material.
As illustrated in the figures, existing video editing and inpainting methods struggle with performing the required local edit, i.e., removing the glasses of the person. Moreover, even when they do remove the glasses, they tend to generate artifacts and unrealistic results, leave glasses remnants, or do not preserve the identity of the person or its original eyelid positions. In contrast, our method successfully removes the glasses while preserving the identity and content of the original video.

We perform a thorough user study to compare our results with the results of TokenFlow, RAVE, ProPainter, 
and our editing pipeline with CN inpaint. In the user study, we do not compare our results to those of Text2Video-Zero, as it only removes the glasses from the input video in 4\% of the test videos, as featured in \cref{fig:results_comp}. 

The user study tests three aspects: glasses-removal, resemblance to the identity in the original video, and realism of the result.
For each video in the survey, we ask the users in which videos are the glasses removed. If both models removed the glasses for that video, we additionally ask which video contains less remnants of glasses, which one looks more realistic, and which one better preserves the identity of the person in the original video.
We ask for remnants of glasses, because we noted that even when other models remove the glasses from the video, they often leave parts of the glasses or their reflections, e.g. as presented in the results of RAVE in \cref{fig:results_comp}.
The user study contained side-by-side video pairs of our results vs. the results of different models for the same input (11 per model, 44 pairs in total), and it was answered by 57 users.
The results of our user study are presented in \cref{fig:user_study}.
The top table shows the percentage of videos for which the users thought each model removed the glasses from the input video. The users thought our model removed the glasses in 98.9\% of the cases, more than any other model.
The bottom table shows the percentage of users that preferred our results over the tested models. 
\cref{fig:user_study} shows that the users preferred our results over all other models, in all the measured aspects: identity preservation, realism, and quality of glasses-removal.

\subsection{Quantitative evaluation}
\label{subsec:quant_eval}

\begin{table}
\caption{\textbf{Quantitative results:} We compare our results to different video inpainting and editing methods, where CN inpaint$^*$ is CN inpaint embedded in our video editing pipeline instead of our model. We present results for two versions of our model --- with and without masks, as elaborated in \cref{subsec:quant_eval}.
We test the fidelity of the results by checking $\Delta G$, the average difference between glasses pixels in the original video frames and the edited one.
We test the identity preservation of the edited video using $ID$ score, and the tradeoff between them using $ID \cdot \Delta G$. Moreover, we test the temporal consistency of the generated videos using the optical flow warp error $E_{warp}$.
% Moreover, we test the similarity of the edited video to the original video using $SSIM$.
}
  \adjustbox{max width=\linewidth}{
  \centering
  \newlength{\mycolumnspace}
\setlength{\mycolumnspace}{1em}
\begin{tabular}{l@{}c@{\hskip\mycolumnspace}c@{\hskip\mycolumnspace}c@{\hskip\mycolumnspace}c}
    \toprule
    Method & $\Delta G_{(\times .01)}\uparrow$ & $ID_{(\times .1)}\uparrow$ & $ID\cdot \Delta G_{(\times .01)}\uparrow$ & $E_{warp (\times 10^{-4})}\downarrow$
    % $SSIM$ $\uparrow$ 
    % & FVD $\downarrow$ 
    \\
    \midrule
    ProPainter~\shortcite{zhou2023propainter} & \textbf{4.0} & 7.5 & \textbf{3.0} & 3.9 \\ %\textbf{0.93} &\textbf{0.24} \\
    FGT~\shortcite{zhang2022flow} & \textbf{4.0} & 7.1 & 2.8 & 4.0 \\
    % FGT \cite{zhang2022flow} & x & x & x & x \\
    % \midrule
    % Masked images & 0.94 & - & 0.27 & - \\
    CN inpaint$^*$~\shortcite{zhang2023adding}                       & 3.0 & 8.0 & 2.3 & 4.1 \\ %0.85 & 0.33 \\
    % \midrule
    TokenFlow\shortcite{geyer2023tokenflow} & 2.0 & 6.9 & 1.4 & 5.7 \\ %0.72 & 0.4 \\
    RAVE~\shortcite{kara2023rave}& 2.0 & 7.4 & 1.5 & 6.8 \\ %0.87 & 0.54 \\
    T2V-Zero~\shortcite{khachatryan2023text2video} & 0.3 & \textbf{9.0} & 0.3 & 8.1 \\ %0.85 & 0.33 \\
    \textbf{V-LASIK (ours)} & \textbf{4.0} & 7.6 & \textbf{3.0} & \textbf{3.6}\\ % 0.84 & 0.42
    \textbf{V-LASIK masked (ours)} & \textbf{4.0} & 7.6 & \textbf{3.0} & 4.2\\
    
    \bottomrule
  \end{tabular}
  }
  \label{tab:quant_vid_edit}
\end{table}

We compare the results of different video editing and inpainting methods to the results of our model in two versions --- with and without masks. As mentioned in \cref{sec:imp} both options exist for our model and there is a tradeoff between them. The masked version better preserves the original video colors, while the non-masked version better removes the glasses and thus is more temporally consistent. These differences are small, hence as shown in \cref{tab:quant_vid_edit}, they only slightly affect the quantitative results.

\textbf{Fidelity:} to test the fidelity of our results, we measure the average difference between the number of pixels with glasses in the original videos vs. the edited ones. To find the pixels that contain glasses in an image, we use a face parser \cite{zheng2022farl} that detects glasses, and apply it to each video frame. Then, we calculate the average difference between the number of pixels with glasses in the original frame and the edited one, normalized by the total number of pixels per frame, and report it as $\Delta G$ in \cref{tab:quant_vid_edit}.
As the $\Delta G$ scores in \cref{tab:quant_vid_edit} show, our method removes the glasses better than all other video editing methods, and is on-par with the inpainting methods ProPainter~\cite{zhou2023propainter} and FGT~\cite{zhang2022flow}. However, as illustrated in \cref{fig:results_comp,fig:results_comp2} and by the results of our user study in \cref{fig:user_study}, although they remove the glasses from most videos, as our method does, they often generate unrealistic results with artifacts around the eyes, which our model does not generate.

\textbf{Identity preservation:} to test identity preservation, we use an ID score ($ID$ in \cref{tab:quant_vid_edit}), which is defined by the average cosine similarity between the face embeddings of the video frames, generated by the face recognition model Arcface \cite{deng2019arcface}. 
We note that neither of these metrics is complete on its own, as an unchanged video would get a perfect $ID$ score, and a random video without glasses would get a very high $\Delta G$ score. 
For example, as Text2Video-Zero \cite{khachatryan2023text2video} does not remove the glasses from most videos, and does not change the videos by much, it achieves a very high $ID$ score.
For this reason, we follow prior work \cite{kara2023rave, cong2024flatten} and also look at $ID \cdot \Delta G$, which quantifies the trade-off between removing the glasses from the video, and remaining faithful to the identity of the person in the original video.
When looking at $ID \cdot \Delta G$, our model achieves the best results, together with ProPainter \cite{zhou2023propainter}. However, as mentioned, the results of our user-study in \cref{fig:user_study} show that our model outperforms ProPainter and all other methods both in terms of glasses-removal, and in terms of realism and identity preservation.

\textbf{Temporal consistency:} to evaluate temporal consistency, we follow previous work \cite{geyer2023tokenflow, zhou2023propainter, lei2020blind, lai2018learning} and calculate the warp error ($E_{warp}$ in \cref{tab:quant_vid_edit}). We use RAFT \cite{teed2020raft} to calculate optical flow between each consecutive pair of frames, warp the former towards the latter, and calculate a masked MSE loss (masking occlusions) between the warped frame and the second frame. As shown in \cref{tab:quant_vid_edit}, the non-masked version of our model produces the most temporally consistent results, compared to all other methods.

\subsection{Generalization}
In this work, we focus on glasses as a case study. We chose glasses as they are a particularly complex task, as explained in \cref{sec:intro}.
However, our method can be generalized to other types of local video editing.

\subsubsection{Stickers} 
\begin{figure}[htpb]
    \centering
    \setlength{\tabcolsep}{0pt}
    \renewcommand{\arraystretch}{0.5}

\begin{adjustbox}{max width=\columnwidth}
    \begin{tabular}{c@{\hskip 0.5em} c c @{\hskip 0.5em} c c @{\hskip 0.5em} c c}

        \raisebox{0.235in}{\rotatebox[origin=t]{90}{Input}} &
        \includegraphics[clip,width=16mm]{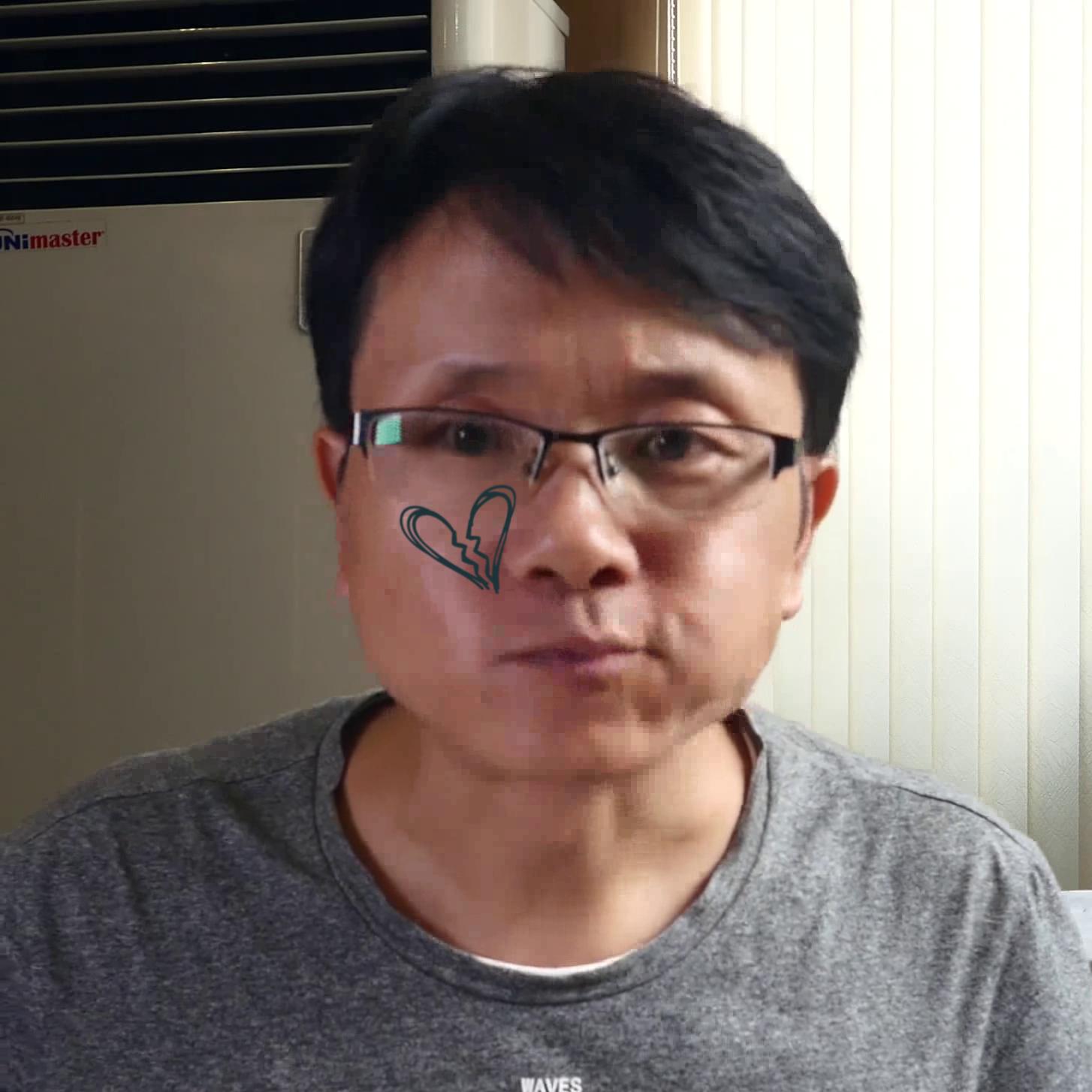} &
        \includegraphics[clip,width=16mm]{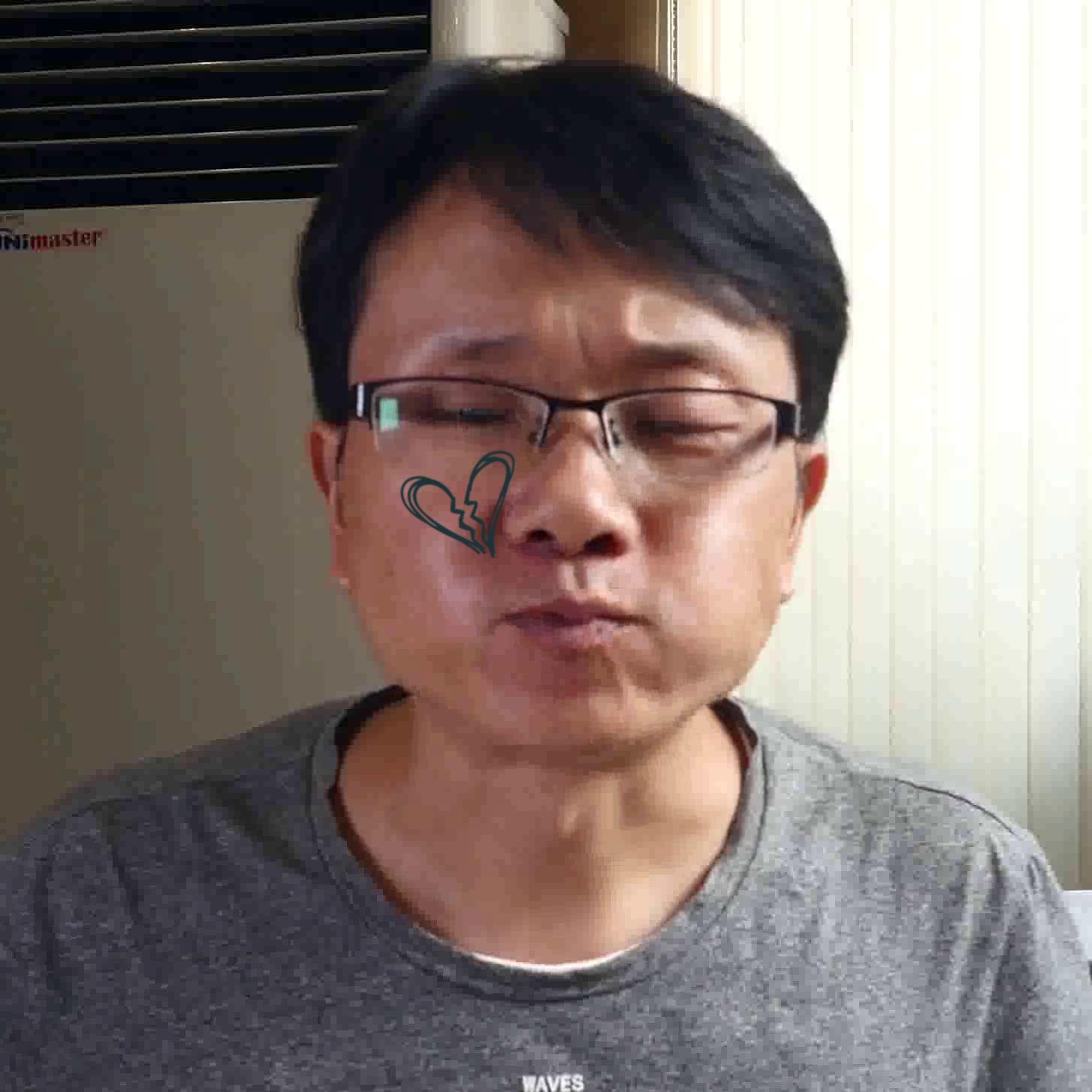} &
        \includegraphics[clip,width=16mm]{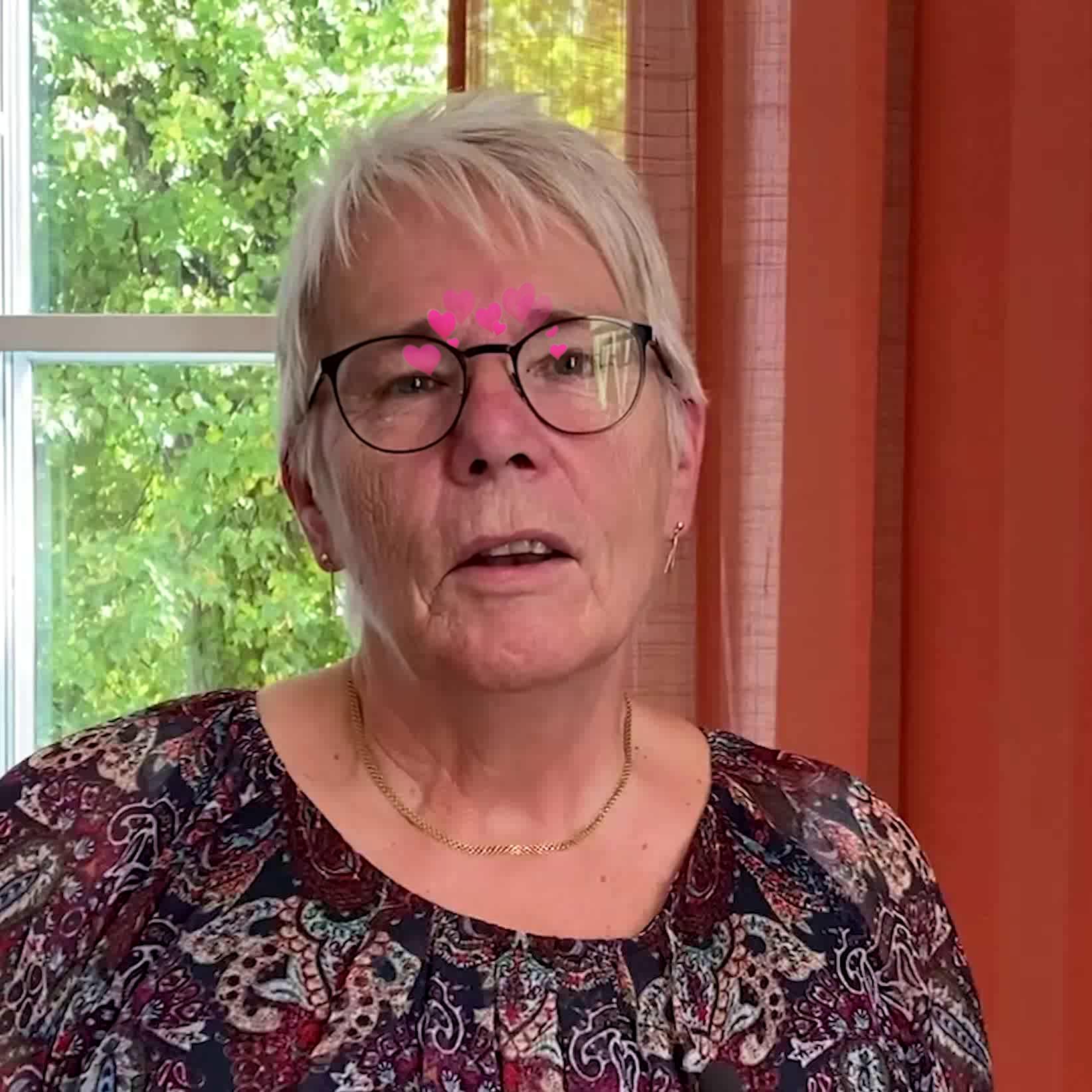} &
        \includegraphics[clip,width=16mm]{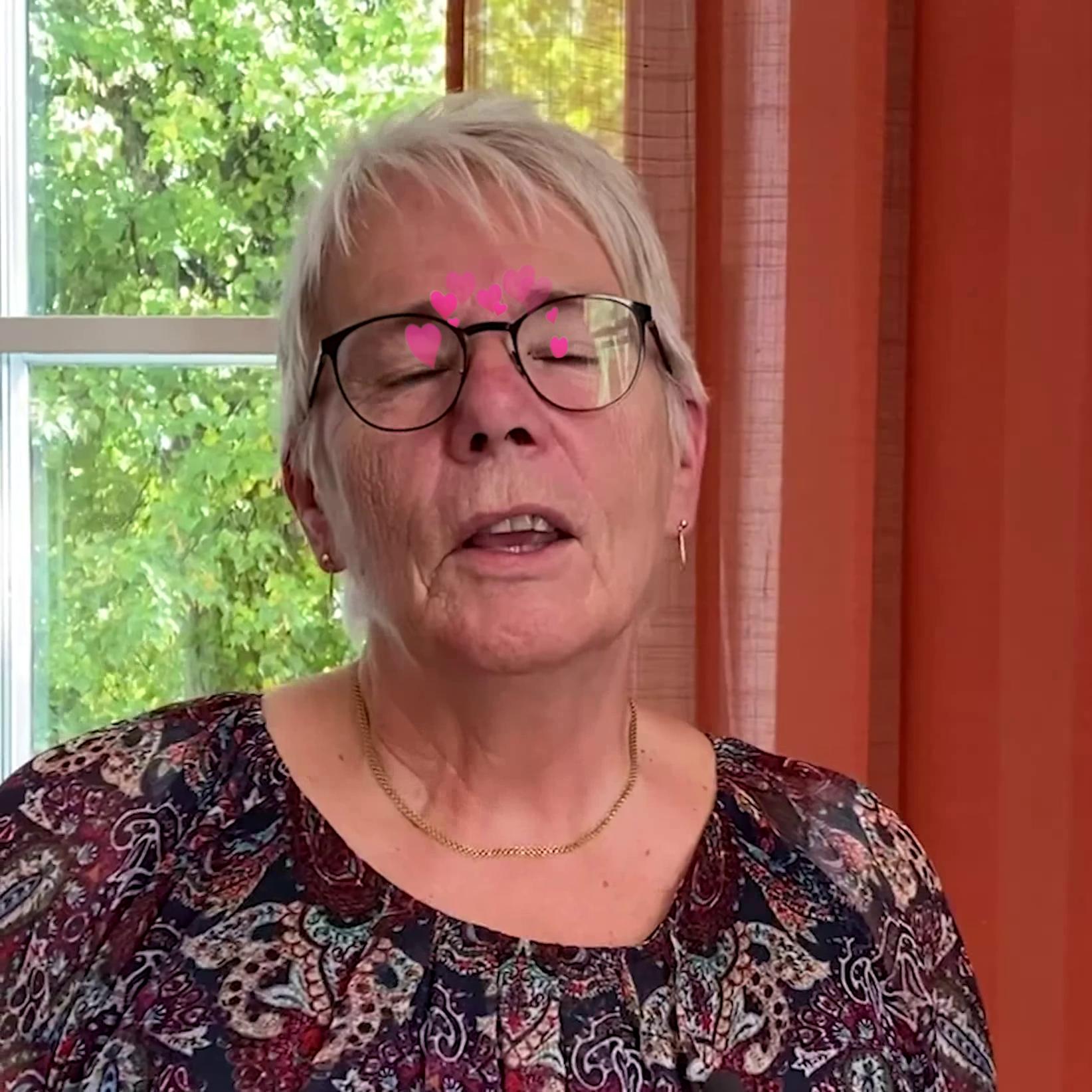} &
        \includegraphics[clip,width=16mm]{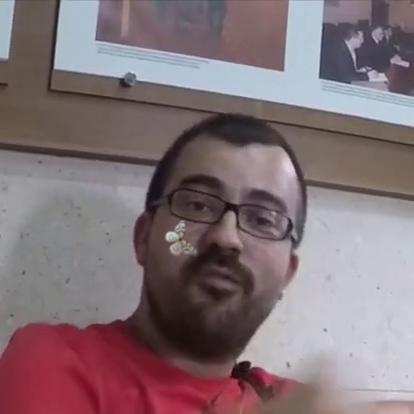} &
        \includegraphics[clip,width=16mm]{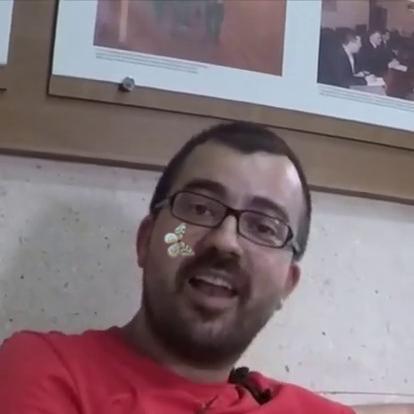} \\

        \raisebox{0.235in}{\rotatebox[origin=t]{90}{Result}} &
        \includegraphics[clip,width=16mm]{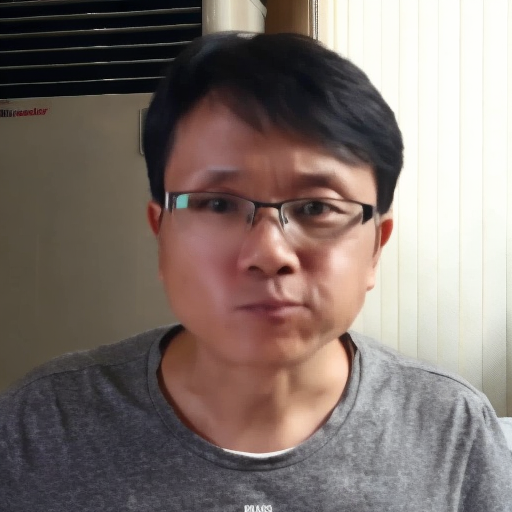} &
        \includegraphics[clip,width=16mm]{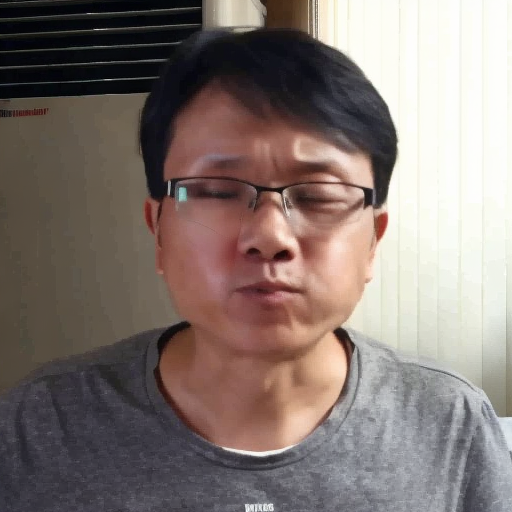} &
        \includegraphics[clip,width=16mm]{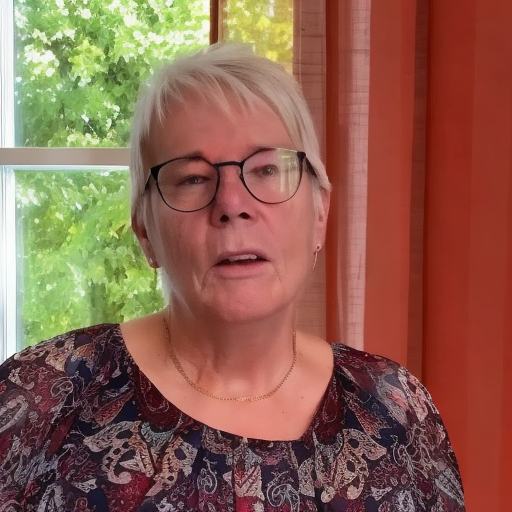} &
        \includegraphics[clip,width=16mm]{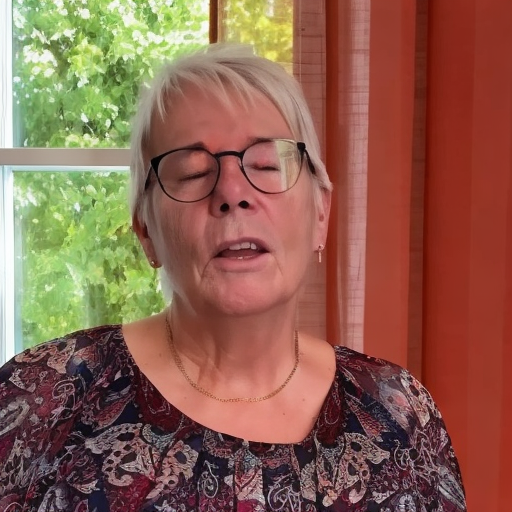} &
        \includegraphics[clip,width=16mm]{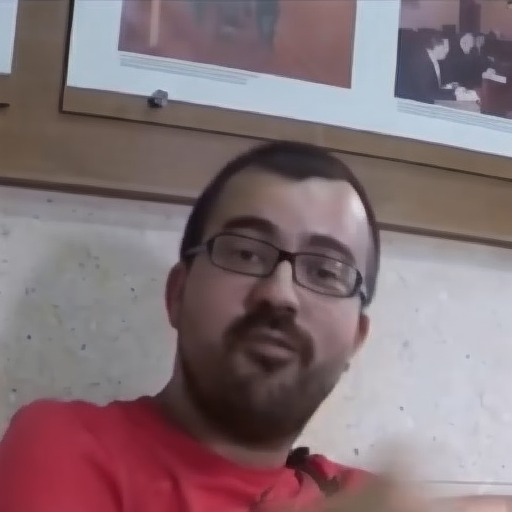} &
        \includegraphics[clip,width=16mm]{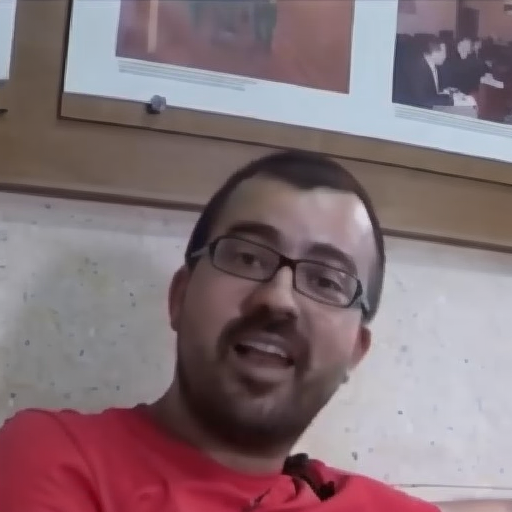} \\

    \end{tabular}
\end{adjustbox}
    \caption{
    \textbf{Stickers results.} These examples show the generalization ability of our method. We apply it to removing facial stickers from videos, and show our model successfully removes different stickers from different locations over the face.
    }
    \label{fig:stickers}
       \Description[]{}  % use this line to please the compiler
\end{figure}
To show our method works for a different use-case, we generate a synthetic dataset of videos by applying stickers from the Stickers dataset \cite{m_Nguyen-etal-CVPR21} on faces from the CelebV-Text \cite{yu2023celebv} dataset, imitating real facial stickers, tattoos or synthetic features added by social-media apps, and show that our method is able to remove them.
For this task, we detect the face in each frame, and estimate its 3D shape using a trained neural network that predicts the coefficients of the Basel 3D face model \cite{gerig2018morphable}. Then, we project and render the sticker texture on the original frame. For each generated clip, we sample one sticker and place it at a random position in the $uv$ coordinate space, which is mapped to the surface of the face. We generate 1274 clips this way, each 40-frames long.

After the dataset is created, we train the same image-to-image model over the stickers dataset and run our video editing pipeline using this model over the test videos. Some results are presented in \cref{fig:stickers}. As shown, our model is able to seamlessly remove the stickers from the videos.
\section{Limitations} 
\label{sec:limitations}
\begin{figure}[htpb]
    \centering
    \setlength{\tabcolsep}{1pt}
    \renewcommand{\arraystretch}{0.5}

\begin{adjustbox}{max width=\linewidth}
    \begin{tabular}{cc@{\hspace{0.5em}}cc@{\hspace{0.5em}}cc}

        Input &
        Output &
        Input &
        Output &
        No MM &
        MM \\

        \includegraphics[clip,width=25mm]{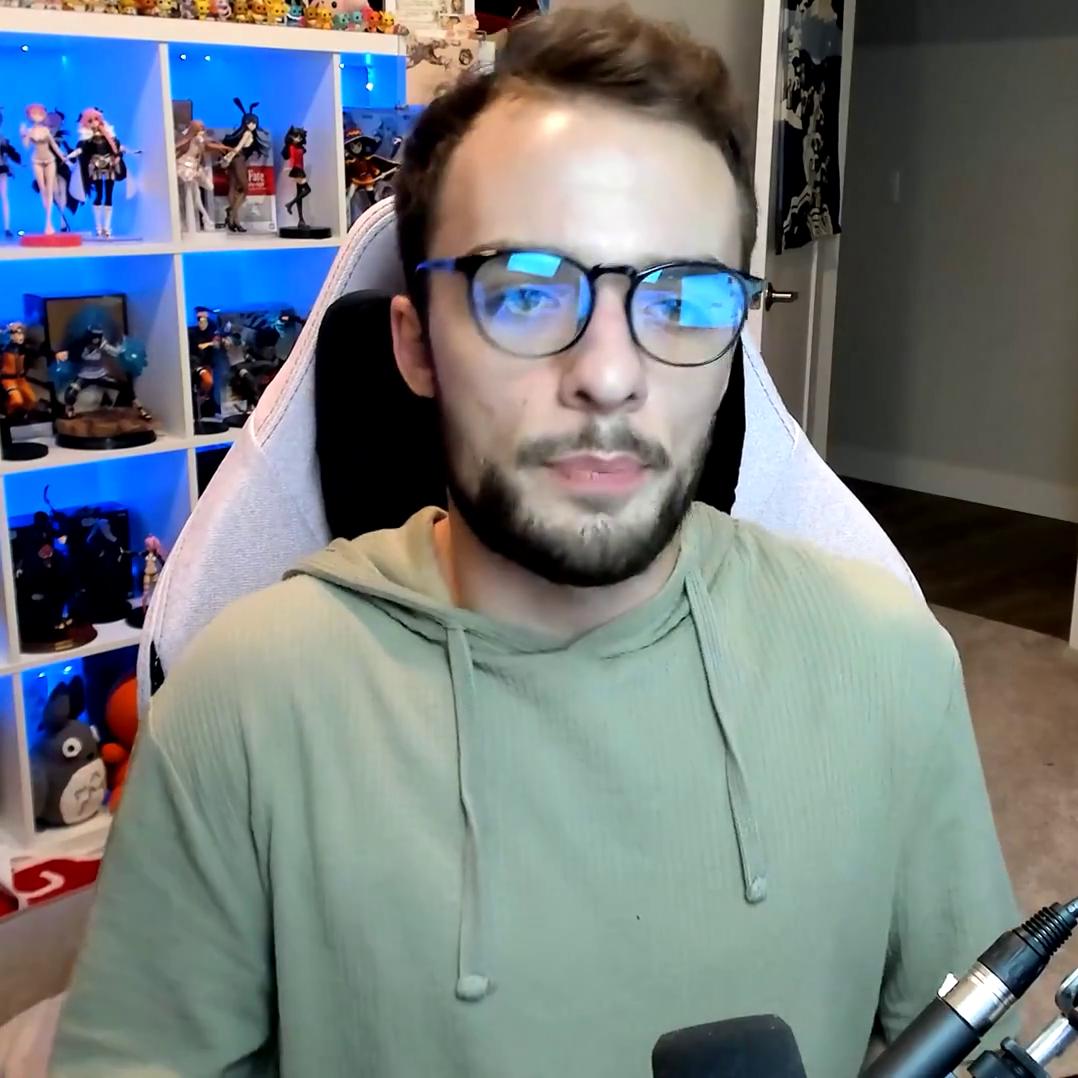} &
        \includegraphics[clip,width=25mm]{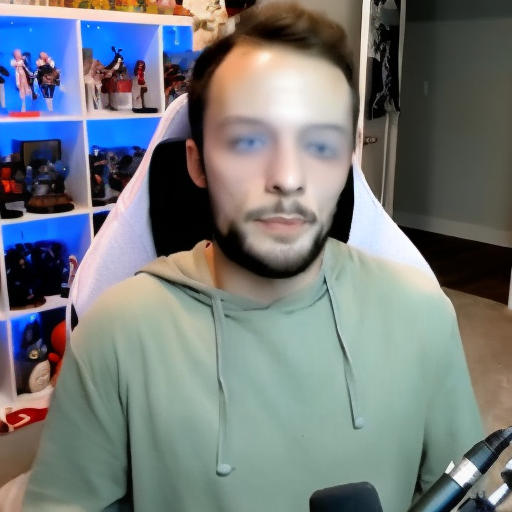} &
        \includegraphics[clip,width=25mm]{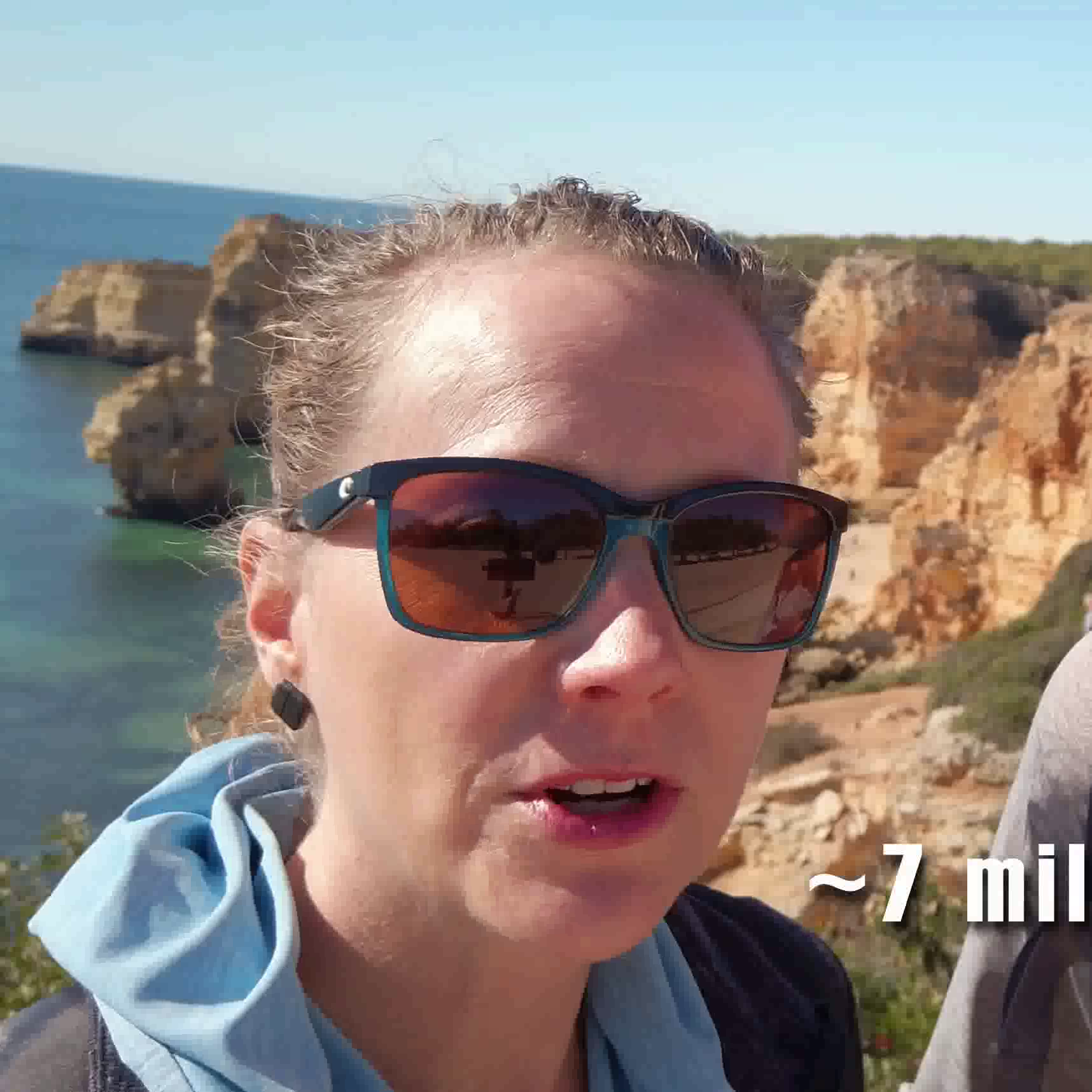} &
        \includegraphics[clip,width=25mm]{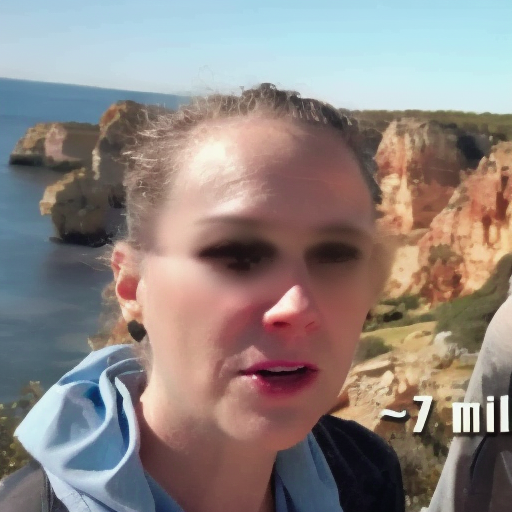} &
        \includegraphics[clip,width=25mm]{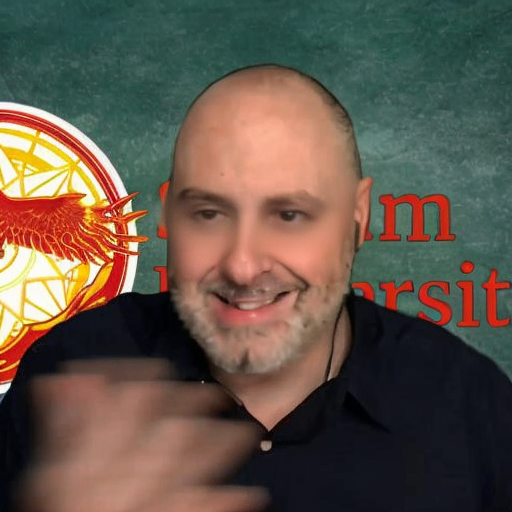} &
        \includegraphics[clip,width=25mm]{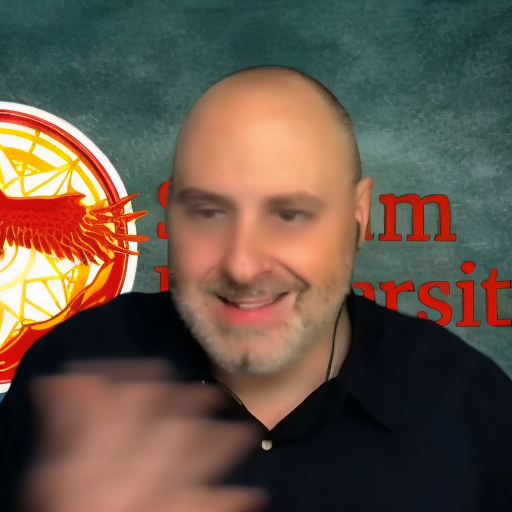}
        \\

    \end{tabular}
\end{adjustbox}
    \caption{
    \textbf{Limitations.} Left: strong reflections. Middle: Dark sunglasses. In these cases, as the eyes are not exposed, the model cannot guess the right eye color. Moreover, our data was not clean and contained sunglasses videos which were treated the same way as eye glasses in the data generation process. Hence, the model generates dark areas instead of eyes for dark sunglasses videos. Right: Eye blurriness that comes from the motion module.
    }
    \label{fig:lim}
       \Description[]{}  % use this line to please the compiler
\end{figure}

Although our method removes glasses reflections in many cases, it struggles with strong reflections, as in the left example in \cref{fig:lim}, where the true eyes are never visible.
Additionally, our method struggles with dark sunglasses, as in the middle example in \cref{fig:lim}, where identifying eye color is difficult.
Moreover, as current glasses detectors do not differ between eye and sun glasses, our dataset was not clean and contained sunglasses videos as well. As we used the same kind of masks for all our data, if the input is a video with sunglasses, our model often generates dark areas where the eyes should be, as is shown the middle example in \cref{fig:lim}. This could be solved by cleaning the data and only training over eye-glasses videos. 
Finally, the motion module tends to smooth the frames to get a more temporally consistent result. As a result, our outputs also tend to be a bit blurry, as shown in the right example in \cref{fig:lim}. As discussed in \cref{sec:imp}, by using only some of the motion layers, we reduce the blurriness to a minimum.

\section{Societal impact} 
\label{sec:societal}

While the goal of this work is to allow editing of owned or licensed videos only, we acknowledge the fact it may be misused to altering videos without consent, and contribute to the spread of misinformation. We condemn such usage, and we are actively working on systems that detect synthetic and edited media~\cite{agarwal2020detecting,knafo2022fakeout,sinitsa2024deep}.
\section{Discussion} 
\label{sec:discussion}

We explored the potential of local video editing through learning from imperfect synthetic data without paired data. Our results surpass existing methods, consistently and realistically removing glasses from videos while preserving the individual's identity.
We focus on the challenging case of removing glasses from videos, however we show our method can also be applied to other local video editing tasks such as removing stickers from faces. We hypothesize it would work similarly for any other local attribute, and we encourage future work to pursue this direction.

\ifanonymous
\else
    \section{Acknowledgements}
\label{ack}

We thank Yotam Nitzan for fruitful discussions.
This project was funded in part by the European Research Council (ERC) under the European Union's Horizon 2020 research and innovation programme, grant agreement No. 802774 (iEXTRACT), and by ISF grant numbers 1574/21, 1337/22.
\fi

% Bibliography
\bibliographystyle{ACM-Reference-Format}
\bibliography{bibiliography}

\end{document}